%% file: main.tex
\documentclass[10pt,twocolumn]{article}
\ifdefined\pdfminorversion\pdfminorversion=7\fi

\usepackage[preprint]{antintlpaper}
\usepackage{xurl}
\usepackage[utf8]{inputenc}
\usepackage{colortbl}
\usepackage[ruled,vlined,linesnumbered]{algorithm2e}
\usepackage{needspace}
\usepackage{float}
\usepackage{placeins}
\usepackage{listings}

\input{macros}

\AntTitle{Verify by Doing: Co-Evolving Agent Execution and Evaluation without External Feedback}
\AntRunningTitle{Verify by Doing}
\AntAuthors{%
  Yitong~Lin\texorpdfstring{\textsuperscript{1,2 *}}{}\and
  Jianda~Chen\texorpdfstring{\textsuperscript{1 *}}{}\and
  Yihao~Hu\texorpdfstring{\textsuperscript{1}}{}\and
  Pan~Wang\texorpdfstring{\textsuperscript{1}}{}\and
  Hang~Wang\texorpdfstring{\textsuperscript{1}}{}\and
  Zhihao~Wen\texorpdfstring{\textsuperscript{1}}{}\and
  Xiaoxue~Ren\texorpdfstring{\textsuperscript{2}}{}\and
  Wei~Wu\texorpdfstring{\textsuperscript{1}}{}}
\AntAffiliations{%
  \textsuperscript{1}Ant International \quad
  \textsuperscript{2}Zhejiang University}
\AntEqualContributions
\AntContact{}
\AntDate{\today}
\AntLinks{%
}
\AntAbstract{\input{sections/abstract}}

\begin{document}
\raggedbottom

\makeanttitle

\input{sections/introduction}
\begin{figure*}[t]
\centering
\includegraphics[width=\textwidth]{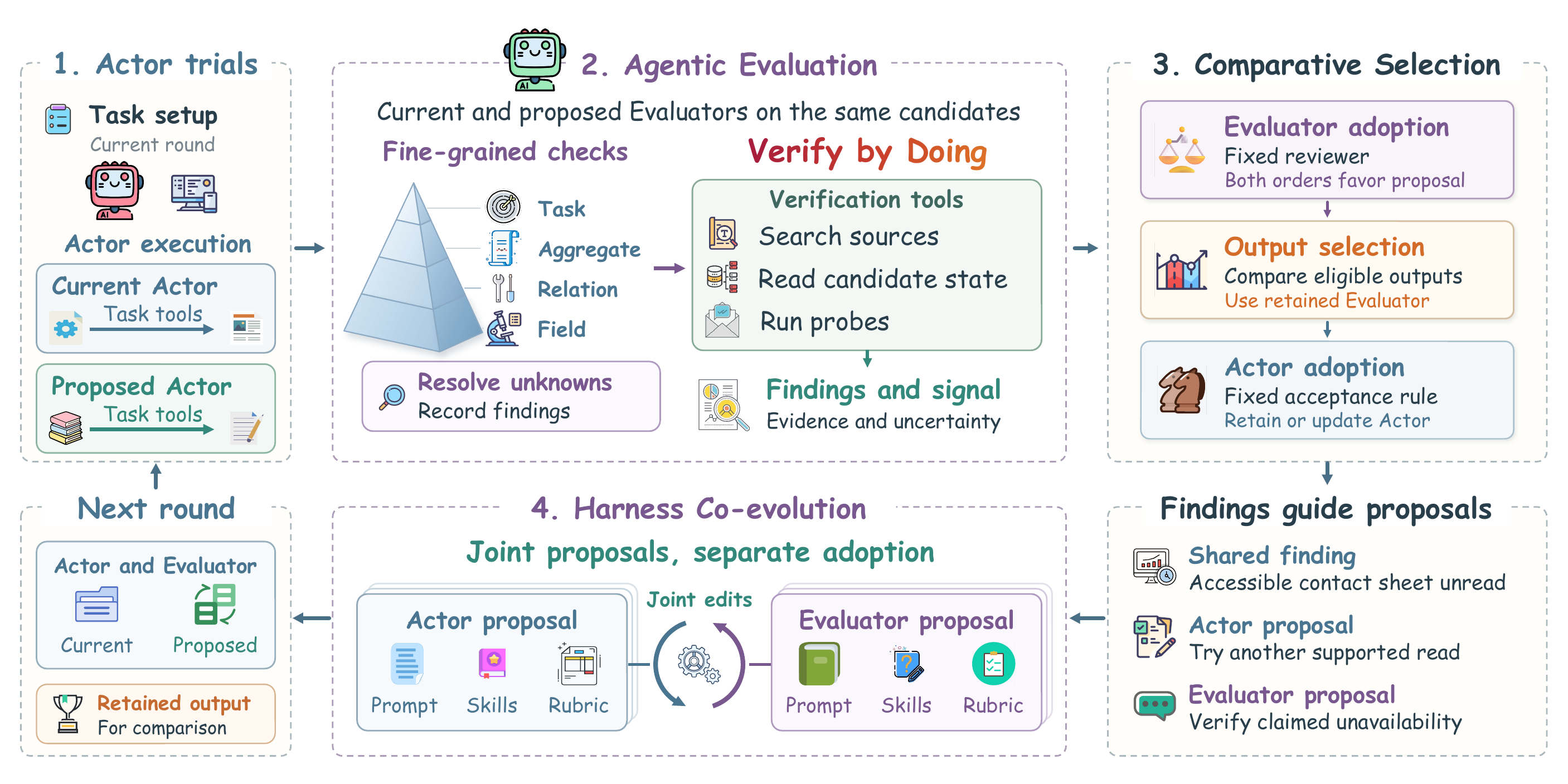}
\caption{\textbf{Verify by Doing overview.} Verification findings provide evidence for the internal signal, guiding output selection and updates to both harnesses. Comparative review and fresh trials support separate adoption decisions for Evaluator and Actor changes. The contact-sheet example connects one finding to changes in both roles.}
\label{fig:overview}
\end{figure*}

\input{sections/related_work}
\input{sections/method}
\input{sections/experiments}
\input{sections/conclusion}

\bibliographystyle{plainnat}
\bibliography{references}

\clearpage
\appendix
\makeatletter
\setlength{\@fpsep}{12pt}
\setlength{\@dblfpsep}{12pt}
\makeatother
\setcounter{dbltopnumber}{3}
\renewcommand{\dbltopfraction}{.9}
\renewcommand{\textfraction}{.08}
\input{sections/appendix}

\end{document}

%% file: macros.tex
\makeatletter
\providecommand{\ant@notice}{}
\providecommand{\AntEqualContributions}{%
  \renewcommand{\ant@notice}{\textsuperscript{*}\,Equal Contributions}}

\newcommand{\vbd@patchemptyfield}[2]{%
  \patchcmd{#1}{\ifstrempty{#2}}{\ifdefempty{#2}}{}{%
    \PackageError{vbd}{Unable to patch optional title metadata}%
      {Review the title metadata compatibility patches in macros.tex.}}%
}
\vbd@patchemptyfield{\ant@authorblock}{\ant@contact}
\vbd@patchemptyfield{\makeanttitle}{\ant@papertype}
\vbd@patchemptyfield{\makeanttitle}{\ant@reportnumber}
\vbd@patchemptyfield{\makeanttitle}{\ant@date}
\vbd@patchemptyfield{\makeanttitle}{\ant@links}
\vbd@patchemptyfield{\makeanttitle}{\ant@keywords}
\makeatother

\definecolor{vbdrow}{RGB}{230,241,250}

\definecolor{deltaup}{RGB}{181,48,55}
\definecolor{deltadown}{RGB}{30,125,88}

\SetKwInput{KwIn}{Input}
\SetKwInput{KwOut}{Output}
\definecolor{algorithmblue}{RGB}{45,78,112}
\newcommand{\Stage}[1]{{\let\nl\relax\textcolor{algorithmblue}{\rule{.55em}{.55em}\enspace\textbf{\scshape #1}}\;}}
\newtcolorbox{answerbox}{
  enhanced,
  colback=black!4,
  frame hidden,
  borderline west={1.5pt}{0pt}{black},
  arc=3pt,
  boxsep=0pt,
  left=6pt,right=6pt,top=4pt,bottom=4pt,
  before skip=4pt,after skip=4pt,
  before upper={\setlength{\parindent}{0pt}}
}
\definecolor{harnessblue}{RGB}{45,78,112}
\definecolor{harnessfill}{RGB}{249,250,252}
\newcommand{\harnessparagraphskip}{4pt}
\newcounter{harnesslisting}
\newenvironment{harnessbox}[2]{%
  \begin{figure*}[p]%
  \refstepcounter{harnesslisting}%
  \begin{tcolorbox}[enhanced,unbreakable,arc=1.2mm,boxrule=.35pt,
    colframe=harnessblue!40,colback=harnessfill,colbacktitle=harnessblue!5,
    coltitle=harnessblue,fonttitle=\fontsize{11}{12.5}\selectfont\bfseries\rmfamily,
    fontupper=\fontsize{10}{12.5}\selectfont\rmfamily,
    boxsep=0pt,left=8pt,right=8pt,top=6pt,bottom=6pt,
    toptitle=5pt,bottomtitle=5pt,before skip=0pt,after skip=0pt,
    title={\strut Listing \theharnesslisting\quad #1\\[0pt]
      {\fontsize{9}{10.5}\selectfont\normalfont\rmfamily\strut #2}},
    title after break={\strut Listing \theharnesslisting\quad #1\hfill
      {\fontsize{9}{10.5}\selectfont\normalfont\rmfamily Continued}}]
  \setlength{\parindent}{0pt}\setlength{\parskip}{\harnessparagraphskip}%
  \setlength{\parfillskip}{0pt plus .5\linewidth}%
  \clubpenalty=10000\widowpenalty=10000%
}{\end{tcolorbox}\end{figure*}}
\newcommand{\componentfield}[2]{%
  \par\noindent{\rmfamily\bfseries\color{harnessblue}#1.}\enspace #2\par%
}
\newcolumntype{L}{>{\raggedright\arraybackslash}X}
\newcolumntype{R}{>{\raggedleft\arraybackslash}X}
\newcolumntype{C}{>{\centering\arraybackslash}X}
\newcolumntype{P}[1]{>{\raggedright\arraybackslash}p{#1}}
\newcommand{\tablestylesetup}{%
  \footnotesize
  \setlength{\tabcolsep}{4pt}%
  \renewcommand{\arraystretch}{1.03}%
}

\newcommand{\tabledivider}{\specialrule{\lightrulewidth}{1.2pt}{1.2pt}}
\newcommand{\tablenote}[1]{%
  \par\vspace{4pt}\noindent
  \parbox[t]{\linewidth}{\footnotesize\raggedright #1}\par
}
\AddToHook{cmd/paragraph/before}{\Needspace{2\baselineskip}}
\AddToHook{cmd/subsection/before}{\Needspace{3\baselineskip}}

%% file: sections/abstract.tex
Agent self-improvement requires feedback about task success, yet external feedback may be unavailable at test time.
Agents must then infer success from observable evidence, and incomplete verification can misguide subsequent updates.
We introduce \emph{Verify by Doing} (VbD), a method that co-evolves Actor and Evaluator harnesses within a single task while keeping model weights fixed.
The Evaluator actively inspects artifacts and environment state with tools; its findings ground an internal signal for output selection and guide updates to both execution and verification procedures.
Proposed Evaluators undergo comparative review on common candidates, while Actor updates require support from their own fresh outputs.
Across AutomationBench, $\tau^3$ Banking, and selected GDPval tasks, VbD achieves the highest overall scores among the compared methods operating without external feedback, with gains of 3.3, 4.1, and 2.3 percentage points over the strongest baselines on the main backbone.
Ablations support the contributions of active verification and Evaluator evolution, and selection from fixed candidate pools shows that improved verification can recover successful outputs without generating new ones.

%% file: sections/introduction.tex
\section{Introduction}
\label{sec:introduction}

Agent self-improvement depends on feedback that identifies successful attempts and guides updates to the procedures that produce them.
At test time, agents can update their workflows or harnesses while keeping model weights fixed~\citep{agrawal2026gepa,lee2026rhi}, or use additional inference to generate and check more candidates~\citep{snell2025testtime,brown2024monkeys}.
Both strategies depend on reliable feedback about whether the task has been completed.
Beyond standardized benchmarks, reference answers and task completion signals may be unavailable, leaving the task request and observations from environment interaction as the main basis for verification.

\begin{figure}[!t]
\centering
\includegraphics[width=\linewidth]{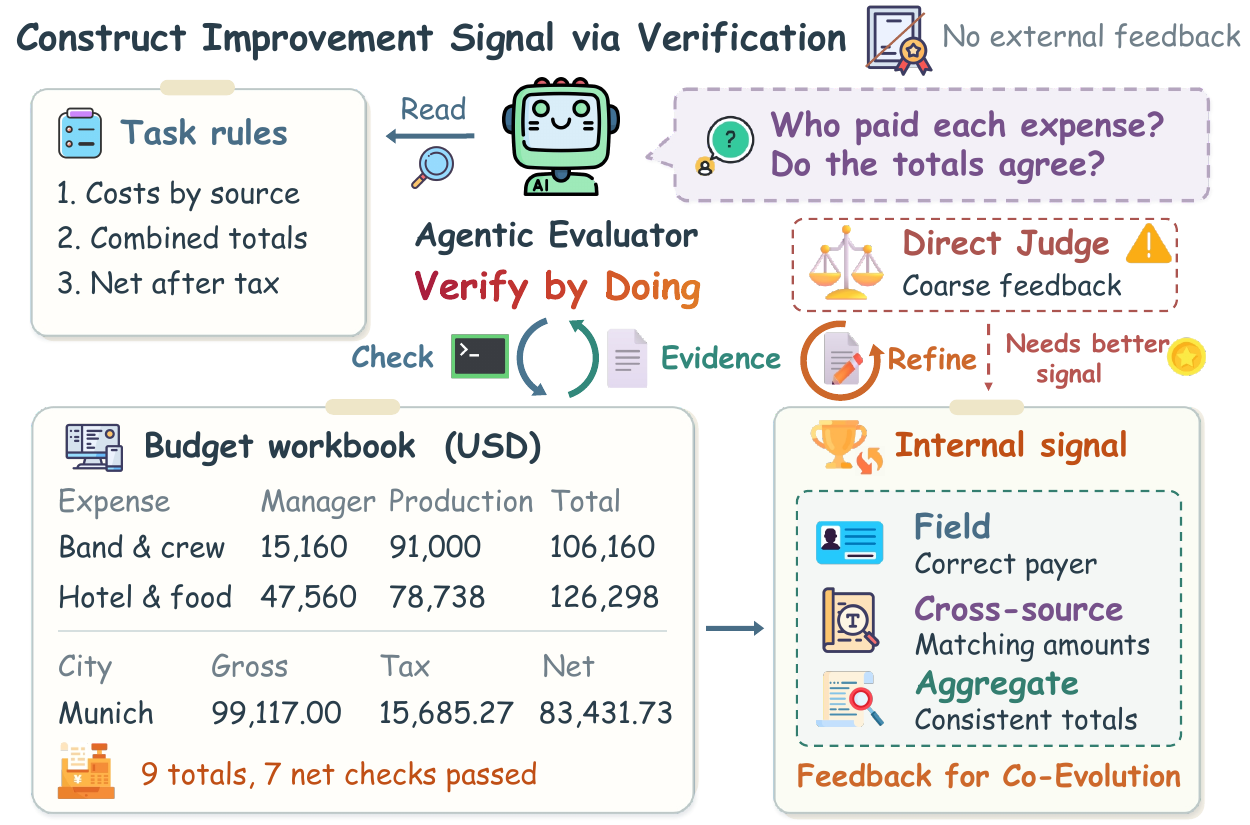}
\caption{\textbf{Active verification of a budget workbook.} The task requires correct payer assignments, consistent expense totals, and after-tax revenue. Initial checks inspect labels and headers; an updated Evaluator skill retrieves expense and revenue cells while the workbook remains selected. Offline recomputation checks expense totals and after-tax revenue from the retrieved cells.}
\label{fig:budget-case}
\end{figure}

This setting creates two coupled challenges: inferring task success from incomplete observations, and improving the agent when the resulting feedback may itself be incomplete. The budget example in Figure~\ref{fig:budget-case} illustrates why both matter: finding expected labels does not establish task completion, while generating more outputs under the same checks can preserve the same blind spots. This motivates active verification and adaptation of the verification procedure.

Existing methods address parts of these challenges. Tool-based critique can acquire additional observations~\citep{gou2024critic}, and retrospective diagnosis and preference history can guide updates to actor procedures~\citep{pan2026rho,lee2026rhi,nie2026tthe}, but they do not update the verification procedure itself.
Incomplete verification can leave blind spots unresolved and misguide later adaptation~\citep{wang2026verificationhorizon}. Red Queen G\"odel Machine jointly evolves agents and evaluators, using held-out ground truth to guide Evaluator replacement~\citep{iacob2026rqgm}. We study within-task adaptation where execution and verification evolve without external feedback from benchmark scores or labeled references.

\label{q:construct}
\label{q:evolve}
\newcommand{\Evaluator}{{Evaluator}}
\newcommand{\Actor}{{Actor}}

We introduce \textbf{Verify by Doing} (\textbf{VbD}), which couples task execution with active verification.
The \textbf{Actor} executes the task through an editable harness of instructions, skills, and working checks, while the \textbf{Evaluator} uses its own harness to inspect artifacts and environment state, recording satisfied, violated, and unresolved requirements.
These findings ground preferences between task results and guide updates to both harnesses. Updating the verification procedure changes how this internal signal is produced, while requirements remain fixed.

After initialization, VbD tests retained and proposed Actor harnesses in the task environment (Figure~\ref{fig:overview}).
Each trial produces a candidate comprising an execution record, resulting state, and response or deliverable.
Proposed Evaluators are reviewed against the retained version on the same candidates, comparing the evidential support and applicable criteria in their reports.
The retained Evaluator's findings guide output selection, while Actor adoption separately requires support for the proposed harness's own fresh output.
An Editor uses the findings, comparison decisions, and task history to propose further updates to either or both harnesses.

VbD makes three contributions:
\begin{itemize}[leftmargin=*, nosep]
\item \textbf{Active verification: verify by doing.} Tool-based inspections gather evidence of task completion from artifacts and environment state.
\item \textbf{Evidence-driven harness co-evolution.} Shared findings guide Actor and Evaluator harness updates within a single task.
\item \textbf{Comparative review and separate retention.} Trials and report comparisons support separate decisions for Evaluator adoption, candidate selection, and Actor retention.
\end{itemize}

Across AutomationBench, $\tau^3$ Banking, and selected GDPval tasks~\citep{patwardhan2025gdpval}, VbD achieves the highest overall scores among the compared methods without external feedback on two backbones.
Ablations support the contributions of active verification and Evaluator evolution, with positive mean gains on all three task sets.
Holding BoN candidates fixed, the retained Evaluator improves pass rate and partial credit over the initial Evaluator, demonstrating better selection from the same candidate outputs.
Case studies connect verification findings to execution improvements and selection of earlier successful candidates, while revealing remaining errors in evidence interpretation.

%% file: sections/related_work.tex
\section{Related Work}
\label{sec:related}
\paragraph{Improving agent procedures.}
Agent improvement methods differ in what they update and how feedback guides those updates. DSPy optimizes modular language-model pipelines, while AFlow searches code-represented workflows~\citep{khattab2024dspy,zhang2025aflow}. Voyager accumulates executable skills through environment feedback~\citep{wang2023voyager}. Self-Refine and Reflexion improve outputs or later behavior~\citep{madaan2023selfrefine,shinn2023reflexion}. GEPA evolves prompts, while DGM, HGM, and Hyperagents edit agent programs or improvement procedures~\citep{agrawal2026gepa,zhang2026dgm,wang2026hgm,zhang2026hyperagents}. Self-Harness accepts edits using regression tests~\citep{zhang2026selfharness}; ModularRSI combines program checks, overfitting review, and execution checks~\citep{wu2026modularrsi}.
Several methods adapt harnesses without labeled validation signals. RHO diagnoses past rollouts and selects harnesses for future tasks using a fixed self-preference ranking procedure~\citep{pan2026rho}. RHI revises a task-specific textual harness from pairwise preference history under a fixed evaluation prompt~\citep{lee2026rhi}; TTHE edits executable harnesses from unlabeled traces~\citep{nie2026tthe}. In contrast, VbD uses verification findings to update both Actor and Evaluator harnesses.

\paragraph{Obtaining evidence for judgment.}
Judging task completion can require observations beyond the Actor's trace. ReAct interleaves reasoning with actions that gather external information~\citep{yao2023react}. CRITIC uses tools to critique outputs, Self-RAG learns retrieval and critique decisions, and LATS combines environment feedback with search~\citep{gou2024critic,asai2023selfrag,zhou2024lats}. Agent-as-a-Judge inspects artifacts against requirements, while VAGEN actively probes graphical interface state~\citep{zhuge2025agentjudge,cui2026vagen}. The Verification Horizon studies informative verification as generation changes; AIChilles searches for workloads exposing regressions~\citep{wang2026verificationhorizon,zhou2026aichilles}. Studies of intrinsic self-correction report failures without external feedback~\citep{huang2024selfcorrect}, while ProCo improves correction through structured condition verification~\citep{wu2024proco}. VbD carries inspection findings into later rounds as proposed changes to both execution and verification.

\paragraph{Co-evolving execution and evaluation.}
Process supervision trains reward models with human feedback on intermediate reasoning steps~\citep{lightman2024verify}. Other training approaches couple generation with learned rewards or feedback on judgments~\citep{yuan2024selfreward,wu2024metarewarding}; related work studies rubric learning and prover--verifier interaction~\citep{ding2026evorubrics,kirchner2024proververifier}. At the program level, RQGM co-evolves agents and evaluators through controlled utility evolution: evaluation remains fixed within each search stage and changes at qualified boundaries~\citep{iacob2026rqgm}. Its shared workspaces also permit edits affecting both roles. Double Ratchet selects evolving metrics with an anchored reference set~\citep{zhang2026doubleratchet}. VbD uses active verification to couple execution and verification updates within a task, without external feedback. Comparative review evaluates evolving feedback procedures before they guide output selection and Actor retention decisions separately.

%% file: sections/method.tex
\section{Problem Setting}
\label{sec:setting}
We study agent improvement within a task $x$, given its request, materials, and permitted tools. Each Actor candidate links an execution record, resulting state, and response or deliverable to its generating harness. Benchmark outcomes, official rubrics, and expert outputs are reserved for offline scoring.

Actor harness $H$ and Evaluator harness $\Phi$ contain instructions, working rubrics, and skills. Rubrics express checks against public requirements: Actor rubrics guide execution and self-checks; Evaluator rubrics guide assessment. Skills specify execution or verification procedures. $H$ also includes persistent check plans, distinct from temporary trial questions. Model weights, task requirements, permissions, editor, selector, and reviewer instructions, eligibility, and acceptance rules remain fixed.

The retained state includes an incumbent candidate $z$, candidate set $\mathcal G$, and history $\mathcal D$. The incumbent need not come from the current Actor. Harnesses and history reset between tasks; the final output is frozen before scoring.

\section{Verify by Doing}
\label{sec:method}
VbD addresses the two challenges in Section~\ref{sec:introduction} by making verification both active and adaptable. Inspection supplies evidence for judging current outputs; the same findings guide changes to how the Actor executes and how the Evaluator checks completion. Because both procedures evolve, VbD reviews verification changes before selection and evaluates Actor changes through fresh trials.

Initialization inspects two attempts from the initial Actor and selects an incumbent without adopting updates. The editor prepares proposals if another round remains. Subsequent rounds follow Figure~\ref{fig:overview}: Actor Trials, Agentic Evaluation, Comparative Selection, and Harness Co-evolution. A tilde denotes a proposal; a superscript $+$ denotes retained state.

\subsection{Actor Trials}
\label{sec:actor-trials}
A verification finding identifies a potential improvement, but the proposed harness must be tested through execution. Following harness optimization~\citep{pan2026rho,lee2026rhi}, Step~1 runs the current Actor $H$ and proposed Actor $\widetilde H$ to produce fresh candidates $a$ and $b$. Both receive the same temporary checks; only the proposal uses persistent edits. This exposes both Actors to the current verification concerns while testing the proposed execution procedure. Temporary checks expire after the round unless reconsidered. Without a proposal, both attempts use $H$.

Each candidate preserves its execution record, resulting state, and response or deliverable for subsequent verification. We append the candidates to $\mathcal G$ and record their generating harnesses. This link matters because an updated Evaluator may select an earlier output: selecting that output provides no trial evidence for adopting the proposed Actor. Fresh trials instead provide execution evidence for the separate retention decision in Section~\ref{sec:trials}.

\subsection{Agentic Evaluation}
\label{sec:vbd}
To judge completion from incomplete observations, Step~2 gathers evidence about the effects of execution. Each Evaluator independently inspects a common pool $\mathcal B$, locating relevant state or artifacts and executing checks for unresolved requirements. Workbook cell reads, for example, expose relationships left untested by header checks (Figure~\ref{fig:budget-case}). Findings take the form
\begin{equation}
\mathcal F_{\Phi}(\mathcal B)
=\{(c_i,q_i,o_i,f_i)\}_{i=1}^{m},
\label{eq:inspection-findings}
\end{equation}
where each of the $m$ findings records a candidate $c_i\in\mathcal B$, proposition $q_i$, observations and sources $o_i$, and scoped assessment $f_i$ with uncertainty. Payer attribution requires linking a value to its source record. Figure~\ref{fig:overview} distinguishes Field, Relation, Aggregate, and Task scopes.

A fixed selector $\Pi$ interprets these findings against the request:
\begin{equation}
d=\Pi\bigl(x,\mathcal B,\mathcal F_{\Phi}(\mathcal B)\bigr).
\label{eq:findings-to-preference}
\end{equation}
Record $d$ names a candidate and distinguishes supported preferences, ties, unresolved judgments, and fallbacks. Its preference supplies the internal signal; the findings guide execution and verification updates. Updating $\Phi$ changes how evidence is acquired and assessed while the task $x$ and selector $\Pi$ stay fixed.

\subsection{Comparative Selection}
\label{sec:trials}
\label{sec:coevolve}
Incomplete checks can favor outputs that miss requirements~\citep{skalse2022rewardgaming,gao2023overoptimization}. Adapting verification can reveal missed requirements but also changes the basis for judging Actor improvement. Step~3 therefore tests the proposed Evaluator on common candidates, uses the retained procedure to select an output, and requires support for the Actor's fresh output before adoption.

\paragraph{Evaluator adoption.}
A fixed reviewer tests whether the proposed verification procedure produces better-supported findings on the same candidates. It compares evidential support, criterion applicability, and counterexamples in the two reports. Both orders are reviewed to reduce sensitivity to position bias~\citep{zheng2023judging}. Let $V_E$ indicate a completed valid trial and $v_1,v_2$ be votes mapped to Evaluator identities:
\begin{equation}
\Phi^{+}=\begin{cases}
\widetilde\Phi, & V_E\land v_1=v_2=\widetilde\Phi,\\
\Phi, & \text{otherwise}.
\end{cases}
\label{eq:evaluator-retention}
\end{equation}
Without a tested proposal, $V_E$ is false and the current Evaluator is retained.

\paragraph{Output selection.}
The retained Evaluator supports selection from the eligible candidates $\mathcal C\subseteq\mathcal G$, with further inspection as needed. Qualification and selection may use different candidate sets (Appendix~\ref{app:comparison-stages}):
\begin{equation}
\begin{aligned}
d^{+}&=\Pi\bigl(x,\mathcal C,\mathcal F_{\Phi^{+}}(\mathcal C)\bigr),\\
z^{+}&=\operatorname{out}(d^{+}).
\end{aligned}
\label{eq:output-selection}
\end{equation}
Here $\operatorname{out}$ returns the named candidate, including a fallback representative. Improved verification can select an earlier successful output even when fresh attempts fail.

\paragraph{Actor retention.}
Actor adoption requires support for the proposal's own fresh candidate. Let $V_A$ denote a valid trial of a changed Actor. Indicators $s_1,s_2$ record strict support in the initial comparison (or candidate-set selection) and the subsequent incumbent comparison, each requiring agreement in both presentation orders:
\begin{equation}
H^{+}=\begin{cases}
\widetilde H, & V_A\land(z^{+}=b)\land s_1\land s_2,\\
H, & \text{otherwise}.
\end{cases}
\label{eq:actor-retention}
\end{equation}
Condition $z^{+}=b$ requires selection of the proposal's own output; $s_1\land s_2$ excludes fallback choices. Selecting an earlier output leaves the Actor unchanged. This rule holds across main experiments and ablations (Appendix~\ref{app:execution-configuration}).

\subsection{Harness Co-evolution}
\label{sec:learning}
\label{sec:meta}
To carry a verification finding into later attempts, Step~4 translates it into a proposed change to execution, verification, or both. The fixed editor $\mathcal L_{\mathrm{edit}}$ specifies an expected effect and a corresponding check for the next trial. Retrieving a supposedly unavailable contact sheet (Figure~\ref{fig:overview}), for example, supplies an Actor retrieval instruction and an Evaluator check for future unavailability claims. The same observation can therefore provide both execution guidance and a verification procedure.

\label{sec:loop}
\label{sec:preference}
\label{sec:evidence-units}
After appending attempts, proposals, and decisions to history $\mathcal D^{+}$, the editor prepares proposals if another round remains:
\begin{equation}
(\widetilde H_{\mathrm{next}},\widetilde\Phi_{\mathrm{next}})
=\mathcal L_{\mathrm{edit}}(x,H^{+},\Phi^{+},\mathcal D^{+}).
\label{eq:harness-proposal}
\end{equation}
History preserves rejected trials and observation sources to inform later proposals~\citep{lee2026rhi}; unchanged components receive no update. Steps~1--3 evaluate these proposals, and their findings guide subsequent edits. Inspection, adaptation, and review connect Actor behavior, Evaluator checks, and the retention of changes. After the final round, VbD returns the selected output (Algorithm~\ref{alg:coupled-iteration}).

%% file: sections/experiments.tex
\section{Experiments}
\label{sec:experiments}
The results support VbD's central premise: execution and its guiding verification procedures benefit from evolving together. Overall gains, fixed-candidate selection, and ablations support this design. Update decisions distinguish output selection from Actor retention; cases and behavioral changes connect verification findings to reusable execution guidance and checks.

\input{tables/main_results}

\subsection{Experimental Setup}
\label{sec:experimental-setup}
\label{sec:reporting}

The three benchmarks require different evidence of task completion. AutomationBench evaluates workflow execution through saved application state on 600 tasks across six domains~\citep{shepard2026automationbench}. $\tau^3$ Banking tests whether dialogue and actions satisfy policy on 97 tasks~\citep{shi2026tauknowledge}. GDPval assesses professional deliverables~\citep{patwardhan2025gdpval}; we use 133 tasks supported by our text and document tools. Appendix~\ref{app:benchmark-scope} gives the GDPval selection criteria.

\input{tables/tau_results}

Here, external feedback refers to benchmark scores or reference labels; the internal signal is the preference inferred from verification evidence (Equation~\ref{eq:findings-to-preference}). We score final selected outputs by pass rate on AutomationBench and Banking, and mean rubric quality on GDPval. Official checks, rubrics, and expert outputs are reserved for offline scoring, except for the reference with external feedback. GDPval's rubric score differs from its official expert pairwise win rate.

The main experiments use DeepSeek-V4-Flash-0731 for all task roles with frozen weights. Offline grading with language models uses GPT-5.6-sol with \texttt{xhigh} reasoning effort. All iterative methods use two Actor executions per round for three rounds, then freeze selected outputs for offline scoring. VbD's main experiments and ablations share comparative Actor acceptance (Equation~\ref{eq:actor-retention}); eligibility and budgets remain fixed within each task. Main tables report mean and sample standard deviation (SD) across three runs. Appendix~\ref{app:seed-reporting} details aggregation and uncertainty.

\subsection{Baselines}
\label{sec:baselines}
Base model makes one attempt (Pass@1); best-of-$N$ (BoN) selects from six candidates using VbD's initial Evaluator. On AutomationBench, self-preference (SP) follows VbD's schedule with both harnesses fixed.

Among adaptive methods, Huxley--G\"odel Machine (HGM) searches Actor policies~\citep{wang2026hgm}, Retrospective Harness Optimization (RHO) diagnoses earlier executions~\citep{pan2026rho}, and Recursive Harness Self-Improvement (RHI) uses preference history~\citep{lee2026rhi}. Red Queen G\"odel Machine (RQGM) co-evolves agents and evaluators~\citep{iacob2026rqgm}; our adaptation preserves staged search and best-belief replacement. Evaluation is fixed within each stage; replacement renews dependent statistics.

All methods include instructions, skills, and working rubrics; evolving baselines reset per task. Permitted task observations replace HGM's benchmark utility and RQGM's labeled anchors (Appendix~\ref{app:prior-methods}). Configurations appear in Tables~\ref{tab:harness-configuration} and~\ref{tab:baseline-harnesses}, and inference costs in Figure~\ref{fig:method-comparison}. The unranked external feedback reference ($\dagger$) runs HGM search under fixed benchmark scoring.

\subsection{RQ1: Task Performance at Test Time}
\label{sec:results}
\label{sec:gdpval-professional}
\emph{Does co-evolving execution and verification improve task performance?} VbD achieves the highest overall scores among the compared methods without external feedback on all three task sets (Tables~\ref{tab:main-results}--\ref{tab:gdpval-results}). The gains span workflow execution, interactions requiring policy compliance, and professional deliverables, supporting the value of improving execution and verification together.

The gains extend to professional deliverables: VbD leads all nine GDPval sectors, with 82.4\% overall rubric quality versus RQGM's 80.1\%. On AutomationBench, selected outputs also satisfy more individual requirements, as shown by higher partial credit (Appendix~\ref{app:partial-credit}).

The evidence differs across tasks, but verification serves a common purpose: inspection findings connect the Actor's actions to observable task requirements. This grounds adaptation in evidence about completion while keeping the task requirements fixed, allowing the same findings to guide both execution and verification.

\input{tables/gdpval_results}
\begin{figure}[t]
\centering
\includegraphics[width=\linewidth]{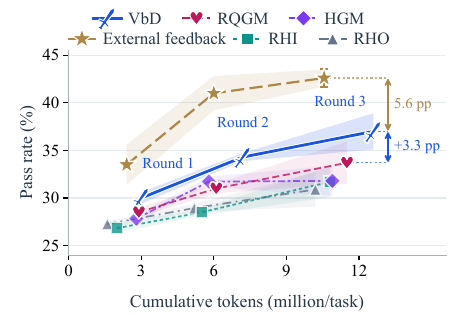}
\caption{\textbf{AutomationBench performance versus cumulative tokens per task.} Shading and bars show SD; arrows compare final-round results. The reference uses external feedback for evolution and selection. Appendix~\ref{app:total-token-scaling} details rounds and costs.}
\label{fig:method-comparison}
\end{figure}

AutomationBench pass rate rises from 30.0\% to 34.3\% and then 37.0\% across rounds (Figure~\ref{fig:method-comparison}). These gains accompany additional inference for verification, editing, review, and selection. VbD's second-round mean already exceeds RQGM's final 33.7\% at a lower cumulative token cost. Each plotted method generates six Actor candidates over its full schedule. VbD finishes 5.6 percentage points below the external feedback reference. Overall gains persist with Qwen3.8-Flash-Next (Appendix~\ref{app:qwen-backbone}).

\begin{answerbox}
\textbf{Answer to RQ1.} Verification is itself a target of test-time improvement. VbD's gains support coupling better task execution with a feedback procedure that learns what to inspect and how to judge success, without external feedback.
\end{answerbox}

\subsection{RQ2: Internal Signal for Co-evolution}
\label{sec:update-gating}
\emph{How does the internal signal support selection and Actor updates?} We assess Actor update decisions within each method's own evolution runs and Evaluator selection on a shared, fixed BoN candidate pool.

We label an Actor proposal as an improvement or regression by comparing its output's offline partial credit with the output it was evaluated against, using valid, non-tied pairs. TP and FP count accepted improvements and regressions; TN counts rejected regressions. We report
\[
\mathrm{TNR}=\frac{\mathrm{TN}}{\mathrm{TN}+\mathrm{FP}},\qquad
\mathrm{Precision}=\frac{\mathrm{TP}}{\mathrm{TP}+\mathrm{FP}}.
\]
\input{figures/quality_control}
VbD rejects 89.3\% of observed regressions compared with 71.2\% for RQGM, while maintaining similar precision (77.6\% versus 77.0\%; Figure~\ref{fig:quality-control}(a)). This advantage of 18.1 percentage points in regression rejection accompanies an update acceptance rate similar to those of HGM and RQGM (Appendix~\ref{app:gating-analysis}).

Incomplete verification can overlook successful outputs already in the candidate pool. To isolate selection, we compare VbD's and RQGM's final Evaluators with BoN's initial Evaluator on six fixed BoN outputs per task (Figure~\ref{fig:quality-control}(b); \mbox{Appendix~\ref{app:evaluator-protocol}}). VbD exceeds both BoN and RQGM, improving pass rate by 4.6 percentage points and partial credit by 2.4 over BoN. Since the available outputs are unchanged, these gains come from selecting better work already in the pool. Marketing088 illustrates the change in feedback: consistently applying the response-scope requirement gives the selector a basis to distinguish similar recorded actions and select an existing passing output (Appendix~\ref{app:independent-pool-case}).

\begin{answerbox}
\textbf{Answer to RQ2.} The internal signal supports both output selection and Actor retention. Improved verification helps identify successful outputs already in the candidate pool, while Actor adoption requires strict support for the proposed harness's own fresh output.
\end{answerbox}

\subsection{RQ3: Active Verification and Evaluator~Evolution}
\label{sec:ablations}
\emph{What do acquiring evidence and adapting verification each contribute?} The variant \emph{without active verification} assesses existing traces without new inspections while retaining harness evolution. The variant \emph{without Evaluator evolution} inspects candidates with fixed Evaluator instructions, rubrics, and skills while the Actor evolves. Tasks, initial harnesses, base model, three-round schedule, and strict Actor acceptance rule are shared.

Both removals lower the mean score on every task set while Actor evolution remains active (Tables~\ref{tab:main-results}--\ref{tab:gdpval-results}). On AutomationBench, removing active verification lowers the mean by 2.7 percentage points, and freezing the Evaluator lowers it by 1.6. Banking and GDPval follow the same pattern.

The ablations distinguish acquiring evidence from retaining verification lessons. A frozen Evaluator still gathers observations, but its instructions, rubrics, and skills remain unchanged. Its lower scores support carrying lessons forward: missed requirements motivate persistent checks, and successful inspections supply reusable procedures. Active verification expands the evidence for current judgments, while Evaluator evolution adapts future checks under fixed task requirements. Actor evolution cannot fully compensate.

\begin{answerbox}
\textbf{Answer to RQ3.} Active verification gathers evidence for current judgments, while Evaluator evolution refines the checks that produce the internal signal. The ablations support both components in improving task performance.
\end{answerbox}

\subsection{RQ4: Co-evolution in Practice}
\label{sec:revision-evidence}
\label{sec:hiring-study}
\label{sec:evolution-case}
\emph{How does a verification finding improve both execution and feedback?} The cases trace how findings become reusable actions and checks through two paths: exposing a missed requirement and discovering an operation the Actor can reuse.

In Sales063, initial attempts omit customer-record updates and violate a reply restriction (Figure~\ref{fig:cases}). Verification accepts an untested claim that the service is unavailable and overlooks the restriction. VbD adds an Actor skill for reply scope and a matching Evaluator criterion; the next attempt completes the updates and satisfies the restriction. HGM and RQGM complete the main actions but retain replies naming excluded attendees. The case shows how a missed requirement becomes guidance for both execution and assessment.

In Marketing015, the Evaluator retrieves a contact sheet through an operation the Actors missed. The successful retrieval becomes an Actor instruction and an Evaluator availability check (Figure~\ref{fig:overview}); distribution then succeeds. The initial Evaluator also recognizes success. Verification supplies more than a verdict here: it reveals an action the Actor can reuse and a check for future claims that a resource is unavailable.

Other cases show why output selection and Actor adoption are separate decisions. Support091 recovers an earlier passing output despite failed fresh attempts. Sales047 selects a passing new output but withholds the Actor update because the final comparison lacks strict support (Appendix~\ref{app:gate-false-rejection}). Verification can still misinterpret evidence: Finance024 treats a file timestamp as a business date and selects a failing output (Appendix~\ref{app:date-failure}).

Actor behavior complements these cases: attempts containing state verification rise from 20.4\% in Round~1 to 25.3\% in Round~3 (Appendix~\ref{app:behavior-change}). The cases trace how findings become Actor checks; the aggregate trend is consistent with this transfer from verification to execution.

\begin{answerbox}
\textbf{Answer to RQ4.} A verification finding can become both a better way to act and a better way to check success. The cases show how shared evidence turns a local discovery into reusable improvements to execution and verification.
\end{answerbox}

%% file: tables/main_results.tex
\begin{table*}[t]
\caption{\textbf{AutomationBench pass rate (\%; mean $\pm$ SD).} Scores of selected outputs across six workflow domains. Bold and underlining mark the best and second-best values; the reference with external feedback ($\dagger$), which uses benchmark scores during evolution, is excluded from ranking.}
\label{tab:main-results}
\centering
\begingroup
\tablestylesetup
\setlength{\tabcolsep}{0pt}
\newcommand{\abresultrows}{%
\toprule
\textbf{Method} & \textbf{Finance} & \textbf{Sales} & \textbf{Marketing} & \textbf{Operations} & \textbf{Support} & \textbf{HR} & \textbf{Overall} \\
\midrule
Base model & $35.3_{\pm 2.1}$ & $20.7_{\pm 2.5}$ & $22.7_{\pm 1.5}$ & $42.3_{\pm 2.1}$ & $10.7_{\pm 1.5}$ & $2.3_{\pm 0.6}$ & $22.3_{\pm 1.6}$ \\
BoN & $\underline{52.7}_{\pm 3.1}$ & $26.7_{\pm 2.1}$ & $39.7_{\pm 2.1}$ & $46.3_{\pm 2.5}$ & $13.7_{\pm 2.1}$ & $8.3_{\pm 1.2}$ & $31.2_{\pm 2.1}$ \\
Self-preference & $41.3_{\pm 2.1}$ & $24.7_{\pm 1.5}$ & $41.3_{\pm 1.2}$ & $38.3_{\pm 2.1}$ & $11.7_{\pm 0.6}$ & $8.7_{\pm 0.6}$ & $27.7_{\pm 1.3}$ \\
\tabledivider
HGM & $44.7_{\pm 2.3}$ & $29.3_{\pm 1.5}$ & $45.3_{\pm 2.1}$ & $47.3_{\pm 2.9}$ & $14.7_{\pm 0.6}$ & $\underline{9.7}_{\pm 1.2}$ & $31.8_{\pm 1.6}$ \\
RHO & $46.3_{\pm 3.2}$ & $29.7_{\pm 0.6}$ & $43.3_{\pm 2.9}$ & $42.7_{\pm 3.5}$ & $14.3_{\pm 1.2}$ & $9.3_{\pm 0.6}$ & $30.9_{\pm 1.8}$ \\
RHI & $45.7_{\pm 2.5}$ & $29.7_{\pm 1.5}$ & $46.3_{\pm 1.2}$ & $44.7_{\pm 3.5}$ & $15.3_{\pm 0.6}$ & $8.7_{\pm 0.6}$ & $31.7_{\pm 1.6}$ \\
\tabledivider
RQGM & $50.7_{\pm 3.1}$ & $30.3_{\pm 2.3}$ & $44.3_{\pm 2.9}$ & $48.3_{\pm 3.2}$ & $\underline{17.7}_{\pm 1.5}$ & $\mathbf{10.7}_{\pm 1.2}$ & $33.7_{\pm 2.2}$ \\
\rowcolor{vbdrow}\textbf{VbD (ours)} & $\mathbf{53.3}_{\pm 3.1}$ & $\mathbf{32.7}_{\pm 2.5}$ & $\mathbf{52.0}_{\pm 2.0}$ & $\mathbf{56.3}_{\pm 3.1}$ & $\mathbf{18.0}_{\pm 2.0}$ & $\underline{9.7}_{\pm 0.6}$ & $\mathbf{37.0}_{\pm 1.9}$ \\
\rowcolor{vbdrow!35}\quad\mbox{w/o active verification} & $49.7_{\pm 2.5}$ & $30.7_{\pm 2.5}$ & $49.3_{\pm 2.1}$ & $52.7_{\pm 2.5}$ & $17.3_{\pm 0.6}$ & $6.3_{\pm 0.6}$ & $34.3_{\pm 1.6}$ \\
\rowcolor{vbdrow!35}\quad\mbox{w/o Evaluator evolution} & $51.7_{\pm 2.5}$ & $\underline{31.7}_{\pm 2.5}$ & $\underline{50.7}_{\pm 1.5}$ & $\underline{54.7}_{\pm 3.5}$ & $17.3_{\pm 1.2}$ & $6.7_{\pm 0.6}$ & $\underline{35.4}_{\pm 1.8}$ \\
\tabledivider
\rowcolor{black!6}External feedback$^{\dagger}$ & $64.7_{\pm 1.5}$ & $40.3_{\pm 1.2}$ & $55.7_{\pm 1.5}$ & $57.3_{\pm 1.2}$ & $26.0_{\pm 1.7}$ & $11.3_{\pm 0.6}$ & $42.6_{\pm 0.9}$ \\
\bottomrule
}
\setbox0=\hbox{\begin{tabular}{l*{7}{r}}\abresultrows\end{tabular}}
\setlength{\tabcolsep}{\dimexpr(\linewidth-\wd0)/16\relax}
\begin{tabular}{l*{7}{r}}
\abresultrows
\end{tabular}
\endgroup
\end{table*}

%% file: tables/tau_results.tex
\begin{table}[t]
\caption{\textbf{$\tau^3$ Banking pass rate (\%; mean $\pm$ SD).} Scores evaluate selected outputs; $\Delta$ is the gain over Base model in percentage points, computed before rounding. The reference with external feedback ($\dagger$) uses benchmark scores during evolution and is unranked.}
\label{tab:tau-results}
\centering
\begingroup
\tablestylesetup
\setlength{\tabcolsep}{0pt}
\newcommand{\tauresultrows}{%
\toprule
\textbf{Method} & \textbf{Pass rate (\%)} & $\Delta$ \\
\midrule
Base model & $35.1_{\pm 3.1}$ & $--$ \\
BoN & $40.5_{\pm 2.6}$ & $5.5$ \\
\tabledivider
HGM & $41.2_{\pm 1.8}$ & $6.2$ \\
RHO & $38.8_{\pm 2.6}$ & $3.8$ \\
RHI & $45.4_{\pm 2.1}$ & $10.3$ \\
\tabledivider
RQGM & $41.2_{\pm 1.0}$ & $6.2$ \\
\rowcolor{vbdrow}\textbf{VbD (ours)} & $\mathbf{49.5}_{\pm 2.1}$ & $14.4$ \\
\rowcolor{vbdrow!35}\quad w/o active verification & $46.0_{\pm 1.6}$ & $11.0$ \\
\rowcolor{vbdrow!35}\quad w/o Evaluator evolution & $\underline{47.1}_{\pm 2.6}$ & $12.0$ \\
\tabledivider
\rowcolor{black!6}External feedback$^{\dagger}$ & $56.4_{\pm 1.6}$ & $21.3$ \\
\bottomrule
}
\setbox0=\hbox{\begin{tabular}{lrr}\tauresultrows\end{tabular}}
\setlength{\tabcolsep}{\dimexpr(\linewidth-\wd0)/6\relax}
\begin{tabular}{lrr}
\tauresultrows
\end{tabular}
\endgroup
\end{table}

%% file: tables/gdpval_results.tex
\begin{table*}[t]
\caption{\textbf{GDPval rubric quality (\%; mean $\pm$ SD).} Selected deliverables from our evaluation subset are assessed offline using official rubrics and expert artifacts. Bold and underlining mark the best and second-best values; the reference with external feedback ($\dagger$) is excluded from ranking.}
\label{tab:gdpval-results}
\centering
\begingroup
\tablestylesetup
\footnotesize
\setlength{\tabcolsep}{0pt}
\newcommand{\gdpresultrows}{%
\toprule
\textbf{Method} & \textbf{Prof} & \textbf{Govt} & \textbf{Manuf} & \textbf{RealEst} & \textbf{Fin} & \textbf{Info} & \textbf{Whsle} & \textbf{Health} & \textbf{Retail} & \textbf{Overall} \\
\midrule
Base model & $65.2_{\pm 1.9}$ & $71.8_{\pm 2.0}$ & $75.0_{\pm 1.9}$ & $73.7_{\pm 2.2}$ & $69.9_{\pm 2.0}$ & $76.3_{\pm 1.9}$ & $70.5_{\pm 2.2}$ & $69.2_{\pm 2.1}$ & $71.3_{\pm 2.0}$ & $70.8_{\pm 1.7}$ \\
BoN & $73.8_{\pm 2.3}$ & $77.3_{\pm 2.6}$ & $79.4_{\pm 2.2}$ & $78.1_{\pm 2.4}$ & $76.5_{\pm 2.4}$ & $80.2_{\pm 2.5}$ & $76.9_{\pm 2.4}$ & $75.4_{\pm 2.5}$ & $76.7_{\pm 2.5}$ & $76.8_{\pm 2.4}$ \\
\tabledivider
HGM & $76.2_{\pm 2.3}$ & $78.3_{\pm 2.6}$ & $80.2_{\pm 2.5}$ & $78.9_{\pm 2.7}$ & $77.9_{\pm 2.5}$ & $79.8_{\pm 2.5}$ & $78.2_{\pm 2.4}$ & $76.8_{\pm 2.6}$ & $77.9_{\pm 2.6}$ & $78.0_{\pm 2.3}$ \\
RHO & $74.0_{\pm 1.5}$ & $76.7_{\pm 1.6}$ & $78.6_{\pm 1.8}$ & $77.5_{\pm 1.9}$ & $76.2_{\pm 1.6}$ & $79.0_{\pm 1.8}$ & $76.6_{\pm 1.6}$ & $75.2_{\pm 1.8}$ & $76.5_{\pm 1.8}$ & $76.4_{\pm 1.7}$ \\
RHI & $76.7_{\pm 2.2}$ & $78.7_{\pm 1.9}$ & $80.2_{\pm 2.0}$ & $79.4_{\pm 2.1}$ & $78.1_{\pm 2.1}$ & $79.6_{\pm 1.9}$ & $78.8_{\pm 2.0}$ & $77.3_{\pm 1.8}$ & $78.3_{\pm 1.9}$ & $78.4_{\pm 1.9}$ \\
\tabledivider
RQGM & $79.5_{\pm 1.8}$ & $80.6_{\pm 1.8}$ & $80.9_{\pm 1.8}$ & $80.4_{\pm 1.7}$ & $79.8_{\pm 1.8}$ & $80.3_{\pm 1.2}$ & $80.4_{\pm 1.9}$ & $79.3_{\pm 1.8}$ & $\underline{80.3}_{\pm 1.8}$ & $80.1_{\pm 1.7}$ \\
\rowcolor{vbdrow}\textbf{VbD} & $\mathbf{81.8}_{\pm 1.6}$ & $\mathbf{82.7}_{\pm 1.6}$ & $\mathbf{83.9}_{\pm 1.5}$ & $\mathbf{83.1}_{\pm 1.4}$ & $\mathbf{82.2}_{\pm 1.5}$ & $\mathbf{82.6}_{\pm 1.5}$ & $\mathbf{82.9}_{\pm 1.7}$ & $\mathbf{81.1}_{\pm 1.3}$ & $\mathbf{81.4}_{\pm 1.5}$ & $\mathbf{82.4}_{\pm 1.4}$ \\
\rowcolor{vbdrow!35}\hspace{2pt}\mbox{w/o active verification} & $79.9_{\pm 2.6}$ & $80.9_{\pm 2.7}$ & $82.0_{\pm 2.6}$ & $81.2_{\pm 2.5}$ & $80.2_{\pm 2.7}$ & $80.6_{\pm 2.4}$ & $80.9_{\pm 2.6}$ & $79.2_{\pm 2.8}$ & $79.2_{\pm 2.6}$ & $80.5_{\pm 2.2}$ \\
\rowcolor{vbdrow!35}\hspace{2pt}\mbox{w/o Evaluator evolution} & $\underline{80.6}_{\pm 2.6}$ & $\underline{81.4}_{\pm 2.9}$ & $\underline{82.4}_{\pm 2.8}$ & $\underline{82.0}_{\pm 3.0}$ & $\underline{80.8}_{\pm 2.6}$ & $\underline{81.4}_{\pm 2.3}$ & $\underline{81.5}_{\pm 2.8}$ & $\underline{79.7}_{\pm 2.9}$ & $79.9_{\pm 2.7}$ & $\underline{81.1}_{\pm 2.4}$ \\
\tabledivider
\rowcolor{black!6}External feedback$^{\dagger}$ & $84.7_{\pm 2.3}$ & $83.1_{\pm 2.5}$ & $85.4_{\pm 2.1}$ & $84.8_{\pm 2.4}$ & $84.3_{\pm 2.2}$ & $85.9_{\pm 2.6}$ & $81.6_{\pm 2.3}$ & $82.7_{\pm 2.1}$ & $81.9_{\pm 2.2}$ & $83.6_{\pm 1.9}$ \\
\bottomrule
}
\setbox0=\hbox{\begin{tabular}{l*{10}{r}}\gdpresultrows\end{tabular}}
\setlength{\tabcolsep}{\dimexpr(\linewidth-\wd0)/22\relax}
\begin{tabular}{l*{10}{r}}
\gdpresultrows
\end{tabular}
\tablenote{Sector labels: Prof = professional services; Govt = government; Manuf = manufacturing; RealEst = real estate; Fin = finance; Info = information; Whsle = wholesale; Health = healthcare; Retail = retail.}
\endgroup
\end{table*}

%% file: figures/quality_control.tex
\begin{figure}[t]
\centering
\includegraphics[width=\linewidth]{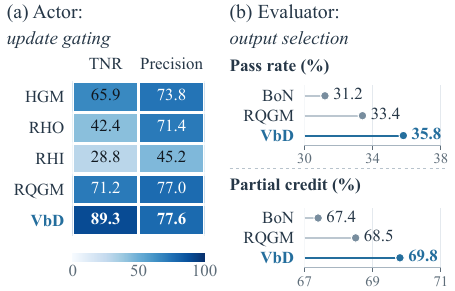}
\caption{\textbf{AutomationBench.} (a) Actor update gating within each method's own evolution runs: TNR and precision. (b) Evaluator selection from shared, fixed BoN pools: pass rate and partial credit.}
\label{fig:quality-control}
\end{figure}

%% file: sections/conclusion.tex
\section{Conclusion}

VbD co-evolves Actor and Evaluator harnesses through active verification, comparative review, and separate retention. Across three task sets and two backbones, it achieves the highest overall scores among the compared methods without external feedback. Ablations and fixed-candidate selection support improving execution alongside the verification procedure that supplies the internal signal.

Together, these findings explain the value of evolving verification alongside task execution. Active verification supplies evidence, co-evolution turns findings into execution guidance and reusable checks, and separate retention distinguishes successful outputs from supported updates. Task requirements and model weights stay fixed; adaptation changes how the agent executes, checks completion, and carries verification findings into subsequent attempts. More broadly, VbD suggests that test-time agent improvement can progress beyond choosing among existing behaviors by evolving the procedures that produce and evaluate feedback.

\section*{Limitations}
\label{sec:limitations}

VbD relies on observable evidence and model-based judgments. The Actor, Evaluator, and reviewer may share errors in interpreting evidence, so active verification and comparative review cannot eliminate incorrect selections or updates~\citep{wang2026verificationhorizon}.

Repeated trials require states or artifacts that can be isolated and inspected. As with replay-based adaptation~\citep{pan2026rho}, irreversible actions and environments that cannot be reset limit direct application. Delayed effects or inaccessible state may also escape verification.

Our experiments evaluate three-round adaptation within individual tasks. Transfer across tasks and sustained improvement over longer evolution remain untested. Additional execution, verification, and harness revision incur inference and tool-use costs; deployment latency also depends on environment response times.

\section*{Ethical Considerations}
VbD studies tool-using agents on benchmark tasks. Errors in execution or verification can lead to unintended messages, incorrect record updates, or misleading deliverables. Harness adaptation retains fixed task requirements and tool permissions. Deployment in real environments should enforce these permissions, protect sensitive information in execution logs and artifacts, and require human oversight for consequential actions. Access to personal or business data should be limited to the authorized task, and use or redistribution of benchmark materials and derived artifacts should respect the applicable licenses and access conditions.

%% file: sections/appendix.tex
\Needspace{20\baselineskip}
\section{Complete Evolution Procedure}
\suppressfloats[t]
\label{app:formal-transitions}
VbD uses verification findings to guide both harnesses, while retaining outputs and updates through separate decisions. Algorithm~\ref{alg:coupled-iteration} specifies how these decisions fit into one co-evolution round after initialization.

The retained state comprises Actor harness $H$, Evaluator harness $\Phi$, incumbent $z$, candidate set $\mathcal G$, and history $\mathcal D$. A tilde denotes a proposal; a superscript $+$ denotes the retained state after the round. Actor state includes persistent check plans. Without an Evaluator proposal, $\Phi^{+}=\Phi$; without an Actor proposal, both fresh attempts use $H$.

\begin{algorithm}[!htb]
\small
\DontPrintSemicolon
\SetAlgoNoLine
\SetAlgoNlRelativeSize{-1}
\caption{One VbD co-evolution round after initialization.}
\label{alg:coupled-iteration}
\KwIn{task $x$; current $H,\Phi$ and proposals $\widetilde H,\widetilde\Phi$; incumbent $z$, candidate set $\mathcal G$, history $\mathcal D$}
\KwOut{retained state, updated history, and next proposals}
\Stage{Actor trials}
Run $H,\widetilde H$ with the same temporary checks (use $\widetilde H=H$ without an Actor proposal) to obtain $a,b$; append them to $\mathcal G$\;
\Stage{Agentic Evaluation}
Form the eligible review pool $\mathcal B$\;
$F\leftarrow\mathcal F_{\Phi}(\mathcal B)$\;
\If{an Evaluator proposal is tested}{
 $\widetilde F\leftarrow\mathcal F_{\widetilde\Phi}(\mathcal B)$\;
}
\Stage{Comparative Selection}
Review $F,\widetilde F$ in both orders if a proposal was tested; retain $\Phi^{+}$ by Equation~\ref{eq:evaluator-retention}\;
Form final comparison pool $\mathcal C$ from eligible candidates; select $z^{+}$ by Equation~\ref{eq:output-selection}\;
Set $H^{+}\leftarrow\widetilde H$ only if its valid trial and selected candidate satisfy both strict comparisons; otherwise set $H^{+}\leftarrow H$ (Equation~\ref{eq:actor-retention})\;
Collect reports, checks, and decisions in $J$\;
Set $\mathcal D^{+}$ to $\mathcal D$ plus $J$, proposals, and attempts\;
\Stage{Harness Co-evolution}
\If{another round remains}{
 Prepare next proposals by Equation~\ref{eq:harness-proposal}\;
}
\Return{retained state, $\mathcal G$, $\mathcal D^{+}$, and any next proposals}\;
\end{algorithm}

\subsection{Evidence, Selection, and Updates}
Both Evaluators inspect a common candidate pool with their respective harnesses fixed during each inquiry. Findings identify the candidate, proposition, observations, and assessment (Equation~\ref{eq:inspection-findings}). The selector interprets these findings against the task requirements, including observed defects and unresolved requirements. A supported finding may establish a defect, such as excessive page count; its consequence for the task determines the preference, rather than the number of supported findings.

Evaluator adoption requires a valid trial and agreement in both review orders. The retained Evaluator supports output selection; Actor adoption additionally requires strict support for the proposed Actor's own output (Equation~\ref{eq:actor-retention}). Fallback selections leave the Actor unchanged.

The editor uses findings and trial history, including rejected proposals, to update either or both harnesses. Each proposal connects an observed gap to an expected change in execution or verification. Task requirements, tool permissions, model weights, controller instructions, and acceptance rules remain fixed. VbD returns the final selected output, then resets its harnesses and history.

\section{Experimental Setup and Results}
\label{app:analysis}
Unless otherwise stated, empirical analyses use DeepSeek-V4-Flash-0731 under the settings in Tables~\ref{tab:harness-configuration}--\ref{tab:model-settings}. We use offline benchmark scores to assess the complete loop, selection by the evolved Evaluator, and Actor retention. Ablations test evidence acquisition and persistent verification updates. Only the external feedback reference uses these scores during evolution (Appendix~\ref{app:oracle-evolution}).

\subsection{Execution Schedule}
\label{app:execution-configuration}
All iterative methods use three rounds with two Actor executions per round, totaling six attempts per task per run. For VbD, Round~1 runs the initial Actor twice to initialize the candidate pool and select an incumbent; no harness update is adopted in this round. Rounds~2--3 test current and proposed Actors using proposals prepared in the preceding round. SP uses the same schedule and budgets with fixed harnesses.

To keep execution and assessment aligned, each candidate records its generating harness and resulting state or deliverable. Evaluators inspect a common pool, and selection uses the retained Evaluator's findings, with further inspection as needed. An earlier candidate remains available through its saved output and state.

\paragraph{Comparison stages and candidate sets.}
\label{app:comparison-stages}
Evaluator qualification uses the common review pool $\mathcal B$; output selection uses the eligible pool $\mathcal C\subseteq\mathcal G$, which may include earlier attempts. Candidate eligibility is fixed throughout evolution and excludes offline task-success labels. Reused evidence retains its candidate identity and observation scope. The main experiments and ablations apply the retention rules in Equations~\ref{eq:evaluator-retention} and~\ref{eq:actor-retention}, with the comparison procedure described in Appendix~\ref{app:formal-transitions}.

After the retention decisions, the editor proposes changes to instructions, rubrics, skills, or persistent check plans and specifies how to test them. Both fresh Actors receive the same temporary checks; only the proposed Actor receives persistent edits. Temporary checks expire after the round unless reconsidered. History records component sources, executed checks, and retention decisions. Changes to persistent Actor check plans count as Actor proposals even when the other harness text is unchanged.

\paragraph{Budget boundaries.}
Operations are bounded by their assigned response limits and remaining task allowance. When no response remains, execution terminates with an error and produces no completed prediction.

\subsection{Tasks, Scoring, and Configuration}
\label{app:harness-configuration}
\label{app:benchmark-scope}
\label{app:ab-fixed-development}
\label{app:tau-fixed-development}
The AutomationBench public set (v1.0.5) contains 600 tasks, with 100 in each of six workflow domains. Banking contains 97 tasks, with success requiring both the correct final database state and the required communication actions. These task sets evaluate completed interactions; GDPval extends the comparison to professional deliverables.

\label{app:gdpval-evaluation}
GDPval selection covers tasks supported by our text and document tools: spreadsheet contents, document structure, measurable formatting, and editable chart data. We select 133 of the 220 public gold tasks, covering nine sectors and 41 occupations.

Selection follows task requirements and reference availability, without filtering by method scores. All methods use the same task IDs and complete rubrics.

\label{app:setup}
\label{app:baseline-protocols}
For evaluation, AutomationBench and $\tau^3$ Banking score each method's final selected output. Pass rate uses the full task denominator; unresolved grades receive no pass credit. Base model makes one attempt, while the other methods submit their selected candidate. All task-local harnesses reset between tasks.

The single-attempt control uses the first BoN candidate, including failed attempts. GPT-5.6-sol with \texttt{xhigh} reasoning effort grades deliverables using official rubrics and expert artifacts. Task scores apply criterion weights and retain penalties; sector scores average their tasks, and overall scores average all 133 tasks within each run. We report the mean and sample SD across three runs. These are local rubric scores, distinct from official expert pairwise win rates. Rubrics and expert artifacts are used for offline scope review and scoring, except in the external feedback reference.

\paragraph{Shared settings and model configurations.}
Table~\ref{tab:harness-configuration} lists the shared execution limits and evolution protocol; Table~\ref{tab:model-settings} gives parameter counts and model-specific inference settings for DeepSeek-V4-Flash-0731 and Qwen3.8-Flash-Next. Full role specifications appear in Appendix~\ref{app:ab-prompts}.

\input{tables/harness_configuration}
\input{tables/qwen_settings}

\begin{figure*}[t]
\centering
\includegraphics[width=\textwidth]{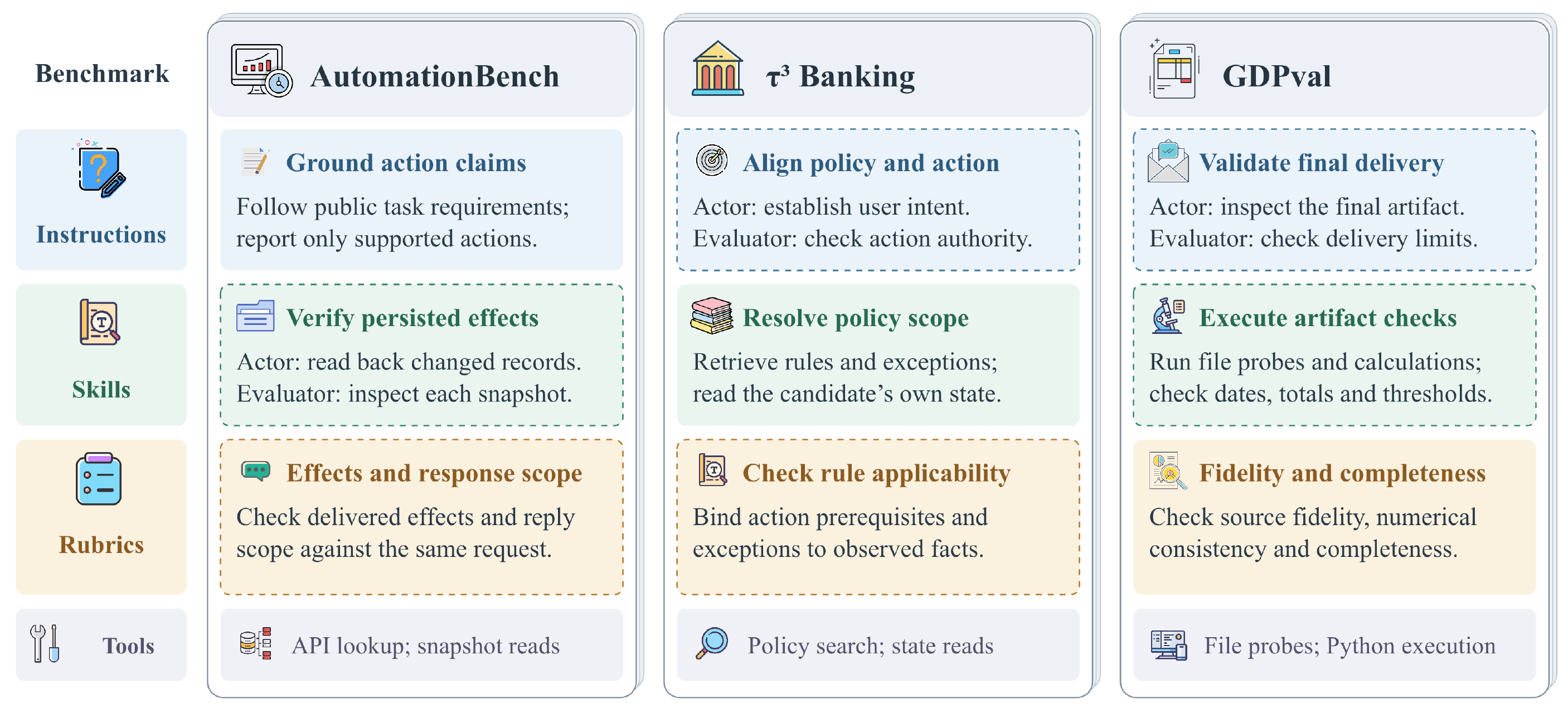}
\caption{\textbf{Verification and harness updates across benchmarks.} A shared evolution loop uses different evidence: application state, policy requirements, or deliverable contents. The examples connect execution requirements to checks of their observable effects.}
\label{fig:benchmark-adapters}
\label{fig:learned-harnesses}
\label{fig:verification-lessons}
\end{figure*}

\label{app:task-interfaces}
The shared loop accesses application state in AutomationBench, policy prerequisites and exceptions in Banking, and file contents and calculations in GDPval (Figure~\ref{fig:benchmark-adapters}). Actor harnesses guide the requested actions; Evaluator harnesses inspect their effects using the permitted task interfaces.

\subsection{Baseline Adaptations}
\label{app:prior-methods}
\input{tables/baseline_harnesses}
Table~\ref{tab:baseline-harnesses} summarizes the evaluated configurations; Table~\ref{tab:prior-methods} compares the original formulations. HGM uses a utility posterior for Actor search and final selection~\citep{wang2026hgm}. Our budget of six attempts limits runtime and token costs, but may reduce HGM and RQGM's tree search to an effectively linear sequence of updates.

RHO originally improves a harness for future tasks by diagnosing grouped rollouts on a coreset of past tasks. It proposes harness changes from these diagnoses and accepts the best proposal only when its mean pairwise self-preference is strictly positive~\citep{pan2026rho}. Our adaptation retains retrospective diagnosis and self-preference-guided Actor updates within the current task. Earlier executions of that task supply the diagnostic history in place of a cross-task coreset, and subsequent attempts test the revised Actor within the six-execution budget. The Evaluator remains fixed throughout adaptation.

RHI originally revises a textual harness using accumulated pairwise preferences between successive executions, with a fixed Evaluator; its stopping criterion aggregates improvement across tasks~\citep{lee2026rhi}. Our adaptation retains task-specific instruction revision from harness and preference history. Each round executes two candidates within the shared six-execution budget, using task-local feedback and a fixed Evaluator. Table~\ref{tab:gating-counts} reports the RHI acceptance statistics.

Both adaptations use the execution and response limits in Table~\ref{tab:harness-configuration}. Harnesses and histories reset between tasks, and benchmark scores remain reserved for offline evaluation.

RQGM additionally changes the Evaluator between search stages. In the original method, evaluation remains fixed within each search stage; replacement uses the highest best-belief score on labeled anchors, optionally augmented with an adversarial objective, and ties retain the incumbent~\citep{iacob2026rqgm}. Our task-local adaptation preserves this structure and replacement rule but estimates utility from permitted task observations. Report comparisons supply evidence for that estimate. Replacement excludes statistics dependent on the displaced Evaluator from subsequent search, while earlier outputs remain available for reassessment.

VbD instead reviews Evaluator proposals within each co-evolution trial and applies the separate retention rules in Equations~\ref{eq:evaluator-retention} and~\ref{eq:actor-retention}.

\paragraph{RQGM utility estimation.}
For each Evaluator--candidate pair, the fixed assessment procedure records whether the report's material findings follow from observed evidence and applicable task requirements. Supported assessments contribute to $S$; evidenced incorrect claims, inapplicable criteria, or material omissions contribute to $F$. Insufficient evidence contributes to neither count. Each pair contributes at most one observation; reassessment updates that observation. A report correctly identifying a failed Actor attempt can therefore receive positive utility. We compute the replacement score as
\begin{equation}
\mathrm{BB}_{\epsilon}(e)=Q_{\epsilon}\!\left[\operatorname{Beta}(1+S_e,1+F_e)\right].
\label{eq:rqgm-proxy-belief}
\end{equation}
Here $Q_{\epsilon}$ denotes the $\epsilon$-quantile, with $\epsilon=0.05$ by default. At a stage boundary, the highest-scoring observed Evaluator replaces the incumbent only on a strict improvement; ties retain the incumbent, and an unobserved alternative cannot displace it.

\input{tables/prior_methods}

\subsection{Evaluator Selection Quality}
\label{app:evaluator-protocol}
\label{app:evaluator-metrics}
\label{sec:selection-evidence}
\label{sec:evaluator-metrics}
To isolate selection quality, we hold Actor outputs fixed. We compare BoN's initial Evaluator with RQGM's and VbD's final retained Evaluators on the same six BoN candidates per task. Each Evaluator uses its own inspection procedure to select one output on each of the 600 AutomationBench tasks. VbD uses the Evaluator retained by Equation~\ref{eq:evaluator-retention}, keeping the initial version if no update is adopted. All selectors use both presentation orders; disagreement is resolved by selecting the smallest original candidate identifier. We report pass rate and partial credit averaged across three runs.

Pass rate is 35.8\% for VbD's Evaluator, 33.4\% for RQGM's Evaluator, and 31.2\% for the BoN baseline (Figure~\ref{fig:quality-control}(b)). Partial credit is 69.8\%, 68.5\%, and 67.4\%, respectively. VbD's evolved Evaluator exceeds BoN by 4.6 percentage points in pass rate and 2.4 in partial credit.

\input{sections/appendix_qwen}

\subsection{Evolution Trajectories and Inference Cost}
\label{app:total-token-scaling}

Figure~\ref{fig:method-comparison} pairs each round's selected-output pass rate with cumulative tokens per task. Cost includes Actor execution, harness editing, evaluation, comparative review, and output selection; it excludes offline grading and the fixed-candidate Evaluator analysis. Each method is plotted at its incurred cost, with final pass rates and SDs from Table~\ref{tab:main-results}. All compared methods generate six Actor candidates per task (BoN-6: six one-shot samples; VbD: two per round over three rounds).

\begin{samepage}
VbD gains 4.3 percentage points from Round~1 to Round~2 and a further 2.7 in Round~3, reaching 37.0\%. Its second-round mean of 34.3\% already exceeds RQGM's final 33.7\% at a lower plotted cumulative cost.
\end{samepage}

\input{sections/appendix_partial_credit}

\input{sections/appendix_oracle}

\subsection{Statistical Reporting}
\label{app:seed-reporting}
\paragraph{Aggregation and uncertainty.}
Main result tables report mean $\pm$ sample SD across three evaluation runs. Overall scores are computed within each run before aggregation; SD is in percentage points, and rounding is applied only for display. Mean differences are reported in percentage points. Behavioral bootstrap intervals summarize variation across tasks using the procedures in Appendix~\ref{app:actor-behavior}.

For GDPval, Tables~\ref{tab:gdpval-results} and~\ref{tab:qwen-gdpval} use the same 133 tasks and sector weights. Within each run, Overall averages task scores, equivalently weighting sector means by sector task counts. Its mean and sample SD are computed across the three run-level Overall scores using unrounded values; each reported entry is then rounded to one decimal place.

\paragraph{Update gating.}
\label{app:gating-analysis}
We assess Actor update decisions over the three-round evolution schedule for each method, using offline partial-credit scores for the recorded proposal--comparison pairs (Table~\ref{tab:gating-counts}). A proposal is positive if its output scores higher than its recorded comparison output, and negative if it scores lower. TP counts accepted improvements, FP counts accepted regressions, and TN counts rejected regressions, using valid, non-tied offline partial-credit pairs. The gate makes its decisions from public evidence; offline scores label the outcomes for this analysis.

For each method, acceptance rate is $N_a/N_p$, where $N_p$ and $N_a$ count all proposed and accepted updates. TNR is \mbox{$\mathrm{TN}/(\mathrm{TN}+\mathrm{FP})$}, and precision is \mbox{$\mathrm{TP}/(\mathrm{TP}+\mathrm{FP})$}; both use valid, non-tied score pairs. All three metrics are percentages.

These measures describe different aspects of retention. Acceptance rate gives its frequency, TNR measures rejection of observed regressions, and precision measures the fraction of improvements among accepted updates. VbD's higher TNR accompanies similar precision and acceptance rates to RQGM. The observed distinction is therefore most pronounced in filtering regressions. This analysis characterizes the complete retention procedure, while the case histories show why an output can be selected without accepting its producing Actor.

\input{tables/gating_counts}

\FloatBarrier
\Needspace{12\baselineskip}
\section{Case Studies of Harness Co-evolution}
\label{app:comparative-cases}

The following recorded cases use DeepSeek-V4-Flash-0731 and trace how verification findings change execution instructions, evaluation criteria, and output selection. They also document updates that are rejected and findings that are misinterpreted, showing where additional evidence does not lead to a better decision.

\subsection{Budget Verification}
\label{app:reward-evolution-case}
The budget task requires correct payer attribution, consistent totals, and tax-adjusted net revenue (Figure~\ref{fig:budget-case}). The initial Evaluator checks labels and headers but leaves the numerical relationships untested. A proposed skill inspects expense and revenue cells through four table probes. Both review orders favor this procedure, while the earlier workbook remains selected.

The figure's nine sum checks and seven net checks are offline recomputations from the returned cells that establish arithmetic consistency for those cells. Payer attribution requires comparing the workbook's assignments with source records.

\subsection{Grant Spending}
\label{app:executed-skill-case}
The grant task classifies spending relative to elapsed project time. An updated skill uses Python to recompute elapsed time and apply the task's thresholds to the recorded spending percentage. For the sampled award, about 16.5\% of the period has elapsed while 99.6\% of the funds are spent, satisfying the rapid-spending condition. This check supports retaining the workbook, although later checks in the episode fail to execute (Listing~\ref{lst:grant-program}).

\subsection{Press Release Distribution}
\label{app:press-release-case}
In Marketing015, both initial Actors fail to read the contact spreadsheet and send no task emails. The Evaluator retrieves the rows through an alternative operation, prompting an Actor retrieval instruction and an Evaluator skill for checking resource availability. In Round~2, the proposed Actor follows the instruction to send the release to three eligible outlets and a summary to the team. Inspection checks a sent release and the summary; both harness updates are adopted, and the passing output remains selected in Round~3. The Actor follows the revised instructions without explicitly loading a skill, and the initial Evaluator also recognizes success. The update therefore supplies a missing retrieval procedure and a check on claims of resource unavailability (Listings~\ref{lst:resource-revision}--\ref{lst:resource-execution}).

\begin{figure*}[t]
\centering
\includegraphics[width=\textwidth]{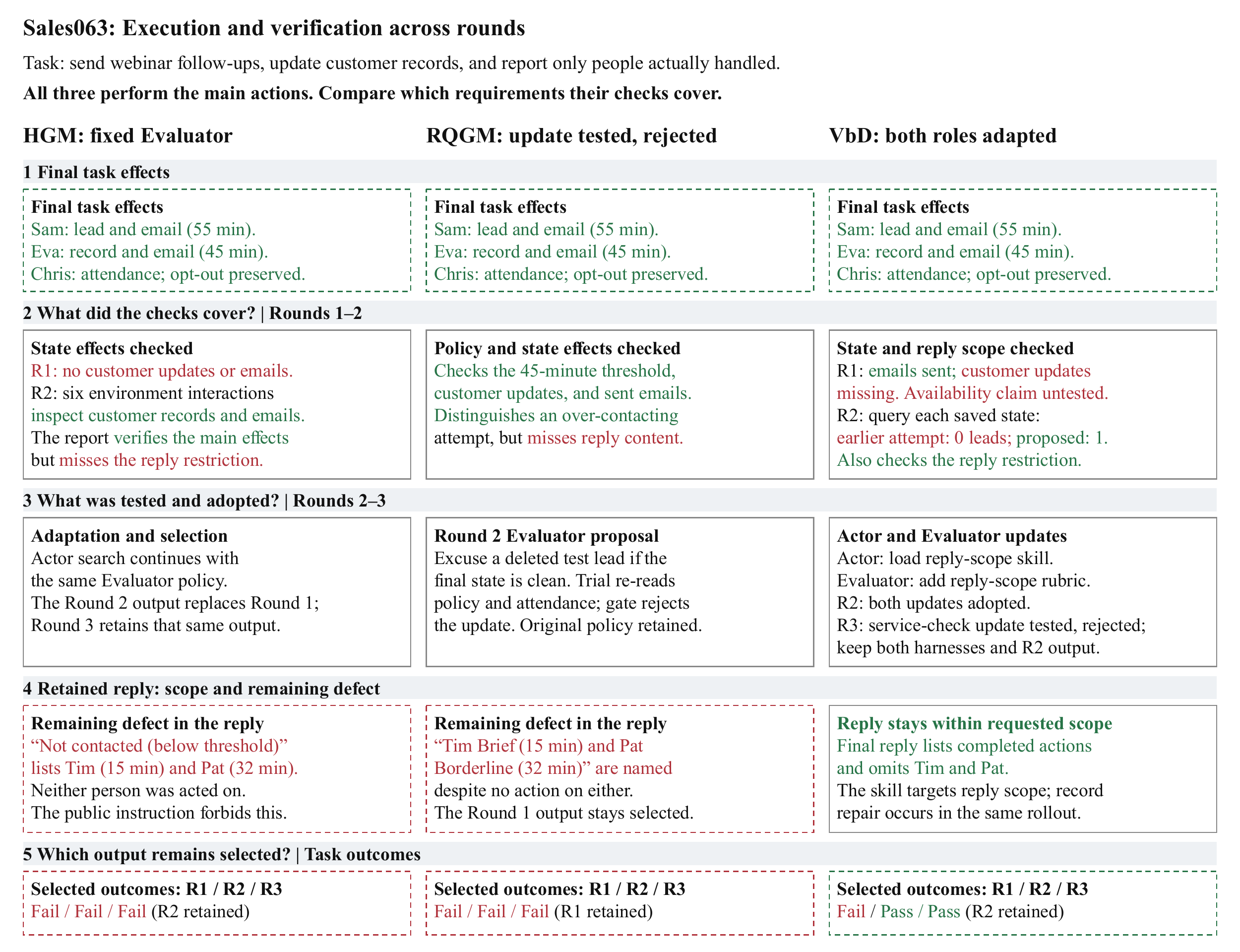}
\caption{\textbf{Sales063: execution and verification across rounds.} Columns trace HGM, RQGM, and VbD using DeepSeek-V4-Flash-0731 on the same AutomationBench task across three rounds. The comparison highlights the missed reply restriction.}
\label{fig:cases}
\end{figure*}

\subsection{Webinar Follow-up}
\label{app:followup-case}
Sales063 requires personalized follow-up emails, updated customer records, and a final reply describing only completed actions. Figure~\ref{fig:cases} and Table~\ref{tab:case-prompt-evolution} trace this task's execution across three rounds.

In Round~1, both initial attempts send follow-up emails but omit updates to customer records and discuss excluded attendees. The selected attempt claims the record service is unavailable without testing it. The Evaluator accepts that explanation and misses the reply restriction. The editor proposes a skill for checking the Actor's final reply and a corresponding Evaluator rubric.

In Round~2, the proposed Actor loads the skill, updates the customer record and attendance, and reports only completed actions. Inspection finds the requested update in its saved state and no matching record in the earlier attempt. The proposed Evaluator checks both the saved state and reply scope, testing the earlier claim of service unavailability against observable results~\citep{wang2026verificationhorizon}. Both updates are accepted, and the selected output passes offline scoring.

In Round~3, a further proposal tests service availability before declaring a limitation. It is not adopted, and the successful Round~2 output remains selected.

HGM and RQGM complete the customer-record updates and emails but retain replies that name excluded attendees. HGM checks the saved records and emails in Round~2 yet retains the reply ``Not contacted (below threshold).'' RQGM distinguishes the updated engagement threshold from a superseded rule and tests an Evaluator update concerning a deleted test lead. That trial produces no findings with stronger evidential support, so RQGM retains its Evaluator and earlier output. VbD's paired updates instead incorporate the missed reply restriction into both execution guidance and assessment.

\begin{table}[t]
\caption{\textbf{Sales063: co-evolution across rounds.} The table follows DeepSeek-V4-Flash-0731 on AutomationBench, with outcomes scored offline.}
\label{tab:case-prompt-evolution}
\centering\tablestylesetup
\begin{tabularx}{\linewidth}{@{}P{.42\linewidth}*{3}{C}@{}}
\toprule
Round & Round~1 & Round~2 & Round~3 \\
\midrule
Actor update adopted & No & Yes & No \\
Evaluator update adopted & -- & Yes & No \\
\tabledivider
Selected output & Round~1 & Round~2 & Round~2 \\
Offline outcome & Fail & Pass & Pass \\
\bottomrule
\end{tabularx}
\end{table}

\begin{figure*}[t]
\centering
\includegraphics[width=\textwidth]{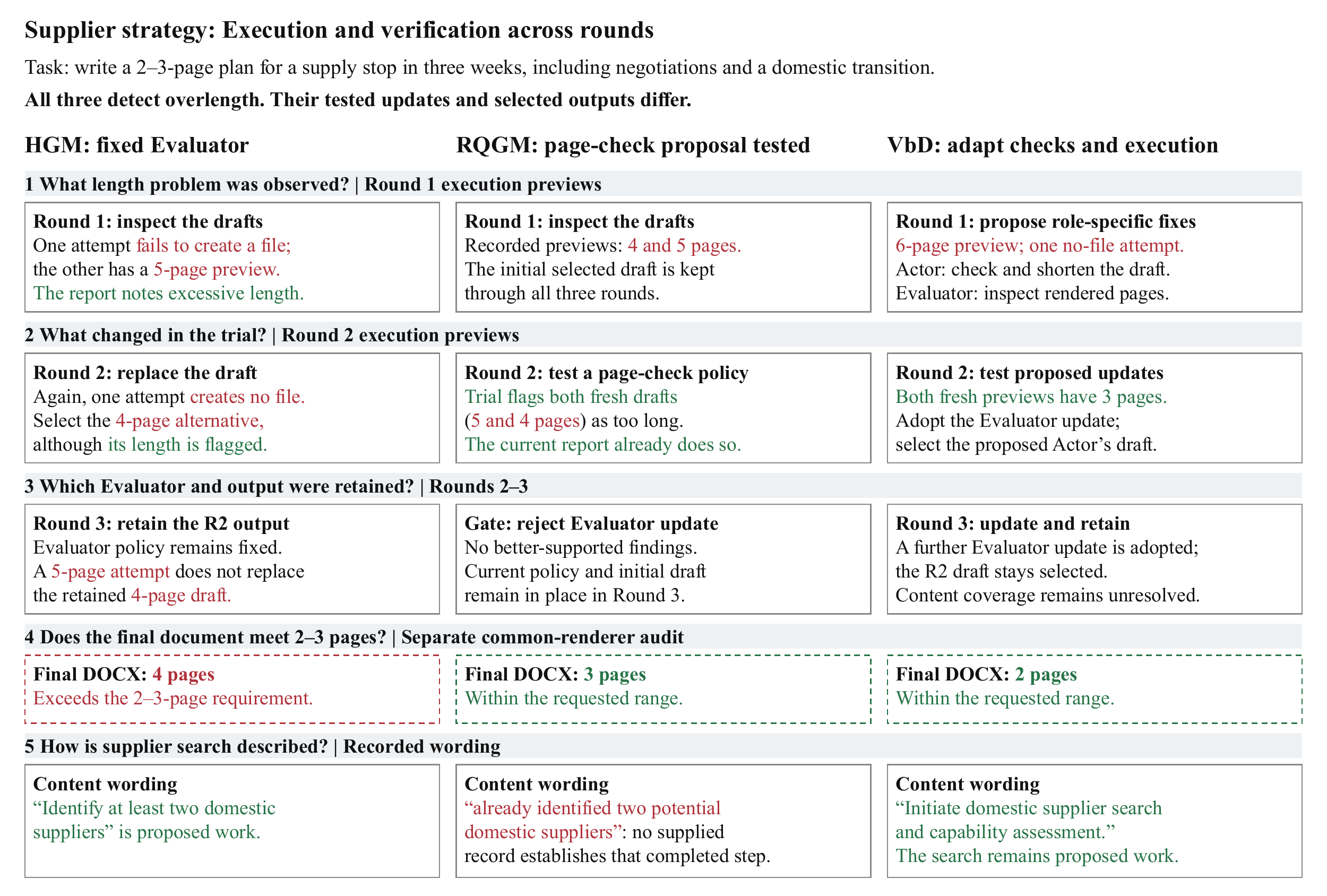}
\caption{\textbf{Supplier strategy: verification and retention across methods and rounds.} For DeepSeek-V4-Flash-0731 on GDPval, previews record page counts during execution; final DOCX lengths use a common renderer. VbD and RQGM meet the requested range.}
\label{fig:supplier-rounds}
\end{figure*}

\subsection{Supplier Strategy}
\label{app:supplier-case}
The supplier task requests a two- to three-page strategy. An initial six-page preview prompts Actor instructions to shorten and check the draft and Evaluator instructions to inspect its rendered length. Both fresh previews show three pages; VbD selects the proposed Actor's draft and adopts the proposed Evaluator.

HGM detects excessive length but retains a four-page draft. RQGM tests a stricter page-limit policy, but both Evaluators already detect the fresh drafts' excess length. It therefore retains its Evaluator and earlier selected draft.

A common-renderer audit finds four pages for HGM, three for RQGM, and two for VbD (Figure~\ref{fig:supplier-rounds}); both RQGM and VbD meet the requested range. RQGM also claims that two domestic suppliers have been identified, whereas the supplied evidence concerns domestic capability. HGM and VbD describe the search as proposed work. VbD retains its draft and adopts another Evaluator update; some content requirements remain unresolved.

\subsection{Evolving Verification of Earlier Attempts}
\label{app:same-pair-case}
In Support091, the Actor remains fixed while revised verification recovers an earlier successful output. The task requires consolidating duplicate tickets, sending notifications, and logging completed actions without mentioning untouched items in the reply. Round~2 finds no supported difference between the compared outputs and retains the initial selection. In Round~3, inspection revisits the same earlier pair and detects excluded-item narration in the initial output. An Evaluator update is adopted, and the passing Round~2 candidate is selected even though both fresh Round~3 attempts fail. Listings~\ref{lst:content-rubric}--\ref{lst:content-execution} document the revised criterion and message reads.

\begin{table}[t]
\caption{\textbf{Support091: updated verification changes output selection.} With DeepSeek-V4-Flash-0731 on AutomationBench, the Actor remains unchanged while output comparisons revisit the same earlier pair. Evaluator qualification comparisons are not shown.}
\label{tab:same-pair}
\centering\tablestylesetup
\begin{tabularx}{\linewidth}{@{}P{.42\linewidth}*{3}{C}@{}}
\toprule
Round & Round~1 & Round~2 & Round~3 \\
\midrule
Actor update adopted & No & No & No \\
Evaluator update adopted & -- & No & Yes \\
\tabledivider
Selected output & Round~1 & Round~1 & Round~2 \\
Offline outcome & Fail & Fail & Pass \\
\bottomrule
\end{tabularx}
\end{table}

In Tables~\ref{tab:case-prompt-evolution} and~\ref{tab:same-pair}, selected-output labels identify the generation round; a dash denotes no Evaluator proposal. Sales063 improves execution and assessment together, whereas Support091 improves selection while retaining the Actor.

\begin{figure*}[t]
\centering
\includegraphics[width=\textwidth]{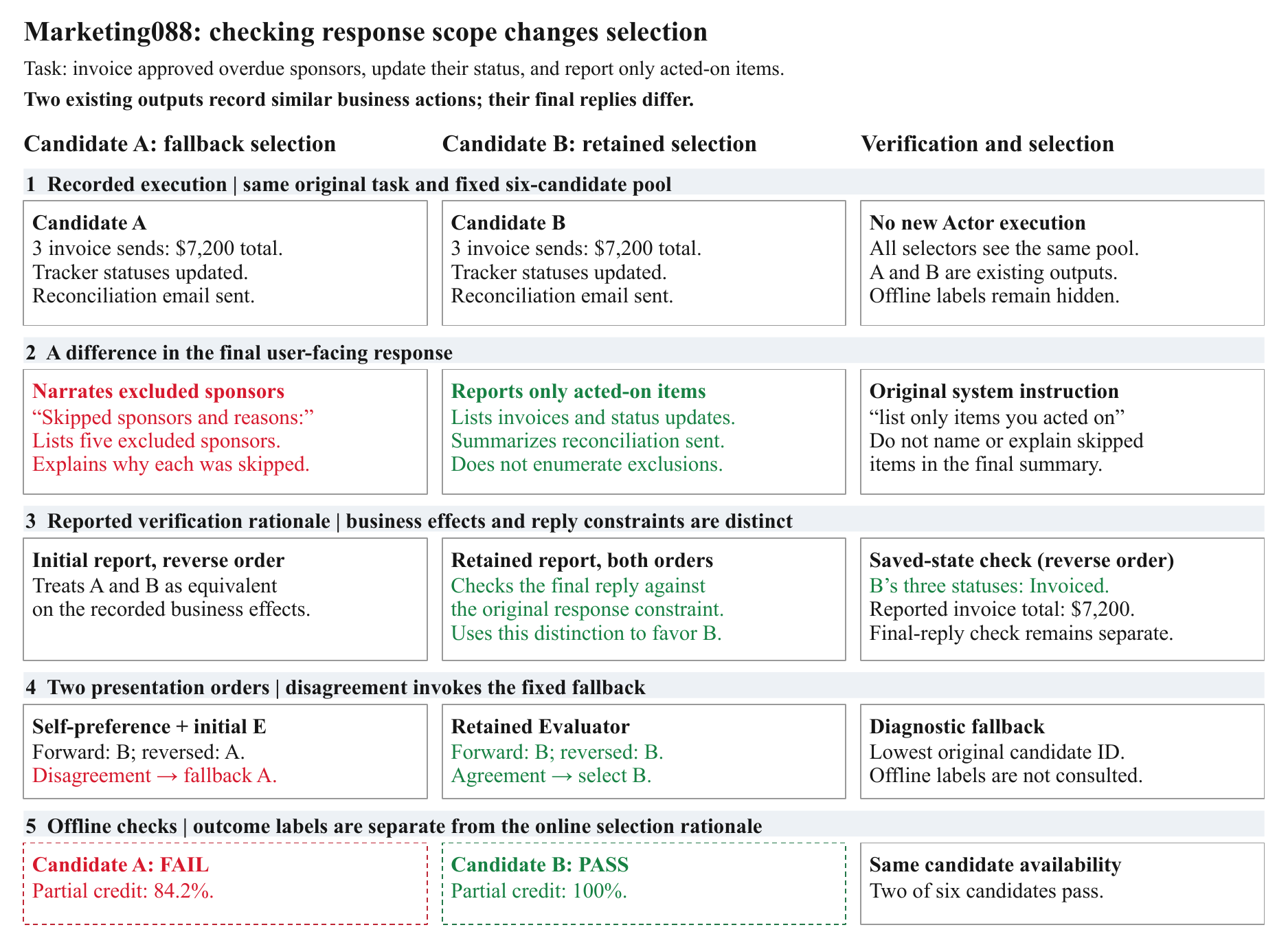}
\caption{\textbf{Marketing088: adapting verification changes the feedback used for selection.} Using DeepSeek-V4-Flash-0731 on AutomationBench with candidate availability fixed, the retained Evaluator distinguishes similar recorded business actions by checking final-response scope. It selects B in both presentation orders; self-preference and the initial Evaluator disagree and fall back to A. Online selection reasons and independent offline outcomes are shown separately.}
\label{fig:independent-pool-case}
\end{figure*}

\subsection{Evolved Verification for Selection}
\label{app:independent-pool-case}
Marketing088 tests whether revised verification can identify a successful output when candidate availability is fixed (Figure~\ref{fig:independent-pool-case}). All selectors receive the same six independently generated BoN outputs. The retained Evaluator selects a passing candidate in both presentation orders, addressing the incomplete-feedback problem described in Section~\ref{sec:introduction}.

The task requires invoicing overdue approved sponsors, applying Finance's processing requirements, and updating the tracker; the final summary must list only acted-on items. Candidates A and B each record three invoice sends totaling \$7,200, status updates, and a reconciliation email. A also lists five skipped sponsors and their exclusion reasons, whereas B reports completed actions without enumerating exclusions.

In the reversed presentation, the initial Evaluator finds no material difference among the most complete candidates on their recorded business effects. The retained Evaluator checks reply scope against the original instruction and distinguishes A from B in both orders. Its reversed-order inquiry also confirms B's three updated tracker statuses. The reply check thus supplies a relevant distinction that the comparison of business effects alone misses.

Self-preference and the initial Evaluator each choose B in the forward order and A in the reversed order; the shared fallback rule resolves the disagreement to A. The retained Evaluator chooses B in both orders. Independent offline checks mark A as failing with 84.2\% partial credit and B as passing with 100\%. These labels are unavailable during selection.

\begin{figure*}[t]
\centering
\includegraphics[width=\textwidth]{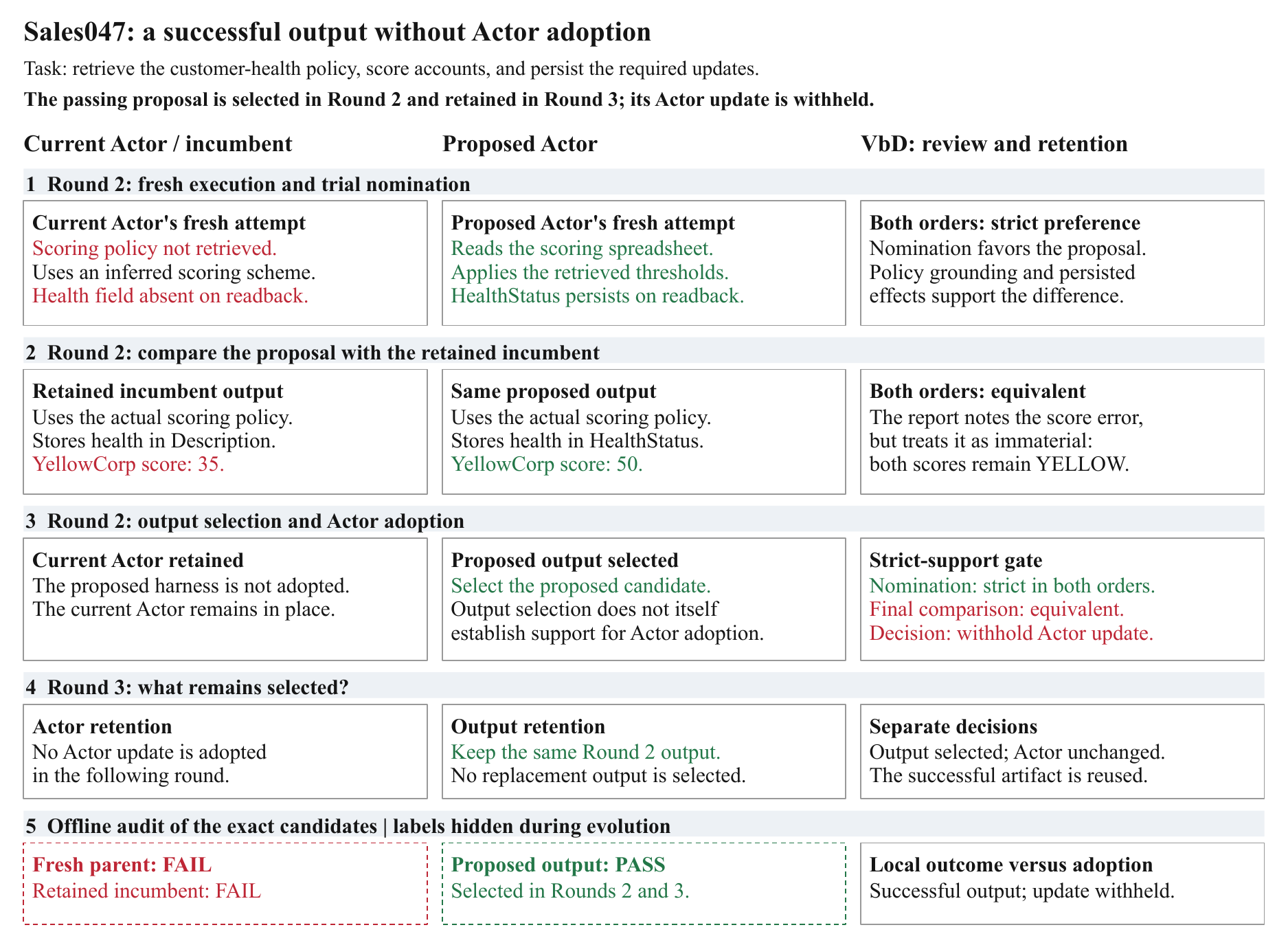}
\caption{\textbf{Sales047: a passing output selected without Actor adoption.} For DeepSeek-V4-Flash-0731 on AutomationBench, the final comparison lacks strict support for retention. Bottom-row outcomes come from offline checks unavailable during selection.}
\label{fig:gate-case}
\end{figure*}

\subsection{Success without Actor Adoption}
\label{app:gate-false-rejection}
In the customer-health scoring task Sales047, a successful output is selected without adopting its Actor update (Figure~\ref{fig:gate-case}). In Round~2, both nomination orders strictly favor the proposal for retrieving the scoring policy and verifying persisted updates. Both final comparisons declare equivalence: the reports acknowledge scores of 50 versus 35 but treat the difference as immaterial because both fall in the same health category. The proposal is selected, but the absence of strict final support blocks Actor adoption. Offline checks mark the proposal as passing and both the incumbent and fresh parent attempt as failing. The same passing output remains selected in Round~3.

\subsection{Failure in Date Interpretation}
\label{app:date-failure}
In Finance024, the selected invoice report passes after Rounds~1--2 but fails after Round~3. The Evaluator substitutes the file's write timestamp for the report's business date, assigns all six invoices to the oldest overdue category, and selects a failing output over the passing incumbent. No harness update is adopted. The failure results from interpreting the observed timestamp as evidence of a different event.

Marketing015 and Sales063 trace findings into changes to both roles, whereas the budget case, Support091, and Marketing088 improve verification or selection of existing work. Sales047 shows that selecting a passing output need not lead to Actor adoption. The supplier and date cases limit what can be concluded from individual checks: a page count establishes length but not content coverage, and a file timestamp does not by itself establish the report's business date.

\begingroup
\setlength{\intextsep}{4pt}
\input{sections/appendix_learned}

\endgroup

\input{sections/appendix_harness_guide}
\clearpage
\input{sections/appendix_prompts}

%% file: tables/harness_configuration.tex
\begin{table*}[!t]
\caption{\textbf{Shared execution and evolution settings for DeepSeek-V4-Flash-0731 and Qwen3.8-Flash-Next.}}
\label{tab:harness-configuration}
\centering
\begingroup
\tablestylesetup
\begin{tabularx}{\textwidth}{P{.25\textwidth}P{.31\textwidth}L}
\toprule
\textbf{Parameter} & \textbf{Value} & \textbf{Scope} \\
\midrule
Backbone assignment & One model per run & Fixed weights; all roles within a task. \\
Context window & 1M tokens & Formatted input plus reserved output. \\
Tool calling & Native & All compared methods. \\
Controller temperature & 0 & Evaluator, editor, and comparative review. \\
Editable components & Instructions, skills, rubrics; persistent check plans & Both harnesses; persistent check plans are Actor state. \\
Evolution rounds & 3 & Initialize in Round~1; test updates in Rounds~2--3. \\
Scheduled Actor trials & 2 per round; 6 per task per run & Follow the three-round schedule. \\
Review orders & 2 & Both must favor the Evaluator proposal. \\
Actor adoption & Comparative acceptance & Main evaluation and ablations; selected new candidate must pass both strict comparisons. \\
State persistence & Within one task & Reset harnesses and history between tasks. \\
Native interfaces & Fixed & Generated checks can change. \\
Evidence during evolution & Public inputs and observations & Official grading remains offline. \\
Offline grading model & GPT-5.6-sol; \texttt{xhigh} & All offline grading with language models. \\
\bottomrule
\end{tabularx}

\vspace{5pt}
\begin{tabularx}{\textwidth}{P{.25\textwidth}CCC}
\toprule
\textbf{Parameter} & \textbf{AutomationBench} & $\boldsymbol{\tau^3}$ \textbf{Banking} & \textbf{GDPval} \\
\midrule
Actor temperature & 1.0 & 1.0 & 0.7 \\
Actor output cap (tokens) & 65,536 & 32,768 & 32,768 \\
Response budget/task & 40 & 40 & 28 \\
Budget scope & Control only & Control only & Generation and control \\
Native trajectory limit & 50 response turns & 200 dialogue steps & --- \\
\bottomrule
\end{tabularx}
\tablenote{Output caps apply per response. Control-only budgets exclude native Actor execution. Banking's user simulator uses temperature 0 and a 4,096-token output cap, with shared task openings across methods. GDPval allows four inspections per evaluation and caps evaluation output at 32,768 tokens per response. The dash denotes no separate native trajectory horizon for GDPval, whose generation is included in its response budget.}
\endgroup
\end{table*}

%% file: tables/qwen_settings.tex
\begin{table*}[!t]
\caption{\textbf{Model configurations.} Shared execution limits are listed in Table~\ref{tab:harness-configuration}.}
\label{tab:model-settings}
\label{tab:qwen-settings}
\centering
\begingroup
\tablestylesetup
\begin{tabularx}{\textwidth}{P{.25\textwidth}CC}
\toprule
\textbf{Parameter} & \textbf{DeepSeek-V4-Flash-0731} & \textbf{Qwen3.8-Flash-Next} \\
\midrule
Backbone parameters & 284B & 125B \\
Activated parameters per token & 13B & 6B \\
Reasoning effort (highest level) & \texttt{max} & \texttt{xhigh} \\
\bottomrule
\end{tabularx}
\tablenote{Reasoning effort applies to task generation and control across all three benchmarks. Parameter counts follow the official model cards~\citep{deepseek2026flashcard,deepseek2026flash0731card,qwen2026flashnextcard} and exclude auxiliary n-gram embedding, MTP, and speculative decoding modules. Qwen3.8-Flash-Next uses YaRN scaling with factor four. Actor top-$p$ is 0.95 for AutomationBench and GDPval and 1.0 for Banking; top-$k$ is 20 for AutomationBench and GDPval. For GDPval, its \texttt{xhigh} setting is subject to an 8,192-token thinking allowance within the output cap. The Banking user simulator has thinking disabled.}
\endgroup
\end{table*}

%% file: tables/baseline_harnesses.tex
\begin{table*}[!t]
\caption{\textbf{Harness configurations used in our comparisons.} All methods include instructions, skills, and working rubrics. Updates and candidate comparisons vary by method; evolving harnesses reset for each task.}
\label{tab:baseline-harnesses}
\centering
\begingroup
\tablestylesetup
\begin{tabularx}{\textwidth}{P{.17\textwidth}P{.24\textwidth}L}
\toprule
Method & Updated components & Adaptation / selection procedure \\
\midrule
Base model & No updates & One Actor attempt. \\
BoN ($N=6$) & No updates & Select among six candidates from one round. \\
Self-preference & No updates & Agentic Evaluation and output retention under VbD's schedule. \\
\tabledivider
HGM & Actor & Posterior-guided Actor policy search with a fixed Evaluator. \\
RHO & Actor & Diagnose prior executions of the current task; guide updates by self-preference with a fixed Evaluator. \\
RHI & Actor instructions & Revise instructions from task-local harness and pairwise preference history with a fixed Evaluator. \\
\tabledivider
RQGM (adapted) & Actor and Evaluator & Staged co-evolution with best-belief Evaluator replacement using permitted task feedback. \\
\rowcolor{vbdrow}\textbf{VbD} & Actor and Evaluator & Coupled Actor--Evaluator adaptation with comparative verification and separate update acceptance and output selection. \\
\bottomrule
\end{tabularx}
\endgroup
\end{table*}

%% file: tables/prior_methods.tex
\begin{table*}[!t]
\caption{\textbf{Comparison with prior methods.} Columns summarize update mechanisms, feedback, and adaptation scope. Our implementations are specified in Table~\ref{tab:baseline-harnesses}.}
\label{tab:prior-methods}
\centering
\begingroup
\tablestylesetup
\setlength{\tabcolsep}{3pt}
\renewcommand{\arraystretch}{1.12}
\newcommand{\priorcomparisonrows}{%
\toprule
\rowcolor{black!8}
\textbf{Method} & \textbf{Adapted components} & \textbf{Feedback} & \textbf{Scope} \\
\midrule
\multicolumn{4}{l}{\textbf{Actor harness optimization}} \\
GEPA~\citep{agrawal2026gepa} & Prompts and instructions & Validation metric + reflection & Across a task set \\
HGM~\citep{wang2026hgm} & Agent code & Benchmark utility & Descendant search \\
RHO~\citep{pan2026rho} & Tools, skills, instructions & Diagnosis + self-preference & Future tasks (offline) \\
RHI~\citep{lee2026rhi} & Agent loop instructions & Pairwise preference history & Within a task \\
TTHE~\citep{nie2026tthe} & Executable harness & Traces + proxy signals & Across batches \\
\addlinespace[3pt]
\multicolumn{4}{l}{\textbf{Actor--Evaluator co-evolution}} \\
RQGM~\citep{iacob2026rqgm} & Actor + Evaluator programs & Utility + held-out labels & Across search stages \\
\rowcolor{vbdrow}\textbf{VbD (ours)} & Actor + Evaluator harnesses & Executed checks; no labels & Within a task; reset \\
\addlinespace[3pt]
\multicolumn{4}{l}{\textbf{Adversarial evaluation}} \\
AIChilles~\citep{zhou2026aichilles} & Candidate test workloads & Differential execution & Program regressions \\
\bottomrule
}
\setlength{\tabcolsep}{0pt}
\setbox0=\hbox{\begin{tabular}{llll}\priorcomparisonrows\end{tabular}}
\setlength{\tabcolsep}{\dimexpr(\textwidth-\wd0)/8\relax}
\begin{tabular}{llll}
\priorcomparisonrows
\end{tabular}
\endgroup
\end{table*}

%% file: sections/appendix_qwen.tex
\subsection{Evaluation Across Model Backbones}
\label{app:qwen-backbone}

With Qwen3.8-Flash-Next, VbD reaches 48.4\% on AutomationBench, 57.4\% on Banking, and 84.7\% rubric quality on GDPval, leading the compared methods without external feedback in all three overall scores (Tables~\ref{tab:qwen-ab}--\ref{tab:qwen-gdpval}). On AutomationBench, it exceeds RQGM by 3.3 percentage points overall and achieves the highest mean score in all six domains among methods without external feedback. It also leads in all nine GDPval sectors. The gains hold across both backbones.

The Qwen3.8-Flash-Next evaluation uses the shared protocol and model settings in Tables~\ref{tab:harness-configuration}--\ref{tab:model-settings}. Baseline schedules follow Section~\ref{sec:baselines}. Offline scoring uses the same task sets and three-run aggregation as the DeepSeek-V4-Flash-0731 evaluation.

\input{tables/qwen_next_ab}
\input{tables/qwen_next_tau3}
\input{tables/qwen_next_gdpval}

%% file: tables/qwen_next_ab.tex
\begin{table*}[t]
\caption{\textbf{Qwen3.8-Flash-Next results on AutomationBench (\%; mean $\pm$ SD).} Values show the mean and sample standard deviation across three runs.}
\label{tab:qwen-ab}
\centering
\begingroup
\tablestylesetup
\footnotesize
\newcommand{\qwencoloredrows}{%
\toprule
Method & Finance & Sales & Marketing & Operations & Support & HR & Overall \\
\midrule
Base model & $39.7_{\pm 2.1}$ & $33.3_{\pm 1.5}$ & $48.7_{\pm 2.5}$ & $33.3_{\pm 2.1}$ & $29.0_{\pm 1.7}$ & $26.7_{\pm 1.5}$ & $35.1_{\pm 1.8}$ \\
BoN & $47.3_{\pm 1.5}$ & $40.7_{\pm 2.1}$ & $58.7_{\pm 2.1}$ & $39.3_{\pm 2.1}$ & $35.3_{\pm 1.5}$ & $32.0_{\pm 1.7}$ & $42.2_{\pm 1.7}$ \\
\tabledivider
HGM & $45.0_{\pm 2.0}$ & $38.3_{\pm 1.5}$ & $59.0_{\pm 2.6}$ & $39.7_{\pm 2.5}$ & $33.0_{\pm 2.0}$ & $30.3_{\pm 1.5}$ & $40.9_{\pm 2.0}$ \\
RHO & $44.7_{\pm 2.1}$ & $38.0_{\pm 2.0}$ & $57.7_{\pm 2.1}$ & $37.3_{\pm 2.1}$ & $34.0_{\pm 1.7}$ & $29.0_{\pm 1.7}$ & $40.1_{\pm 1.7}$ \\
RHI & $49.0_{\pm 2.0}$ & $41.0_{\pm 2.0}$ & $61.0_{\pm 2.6}$ & $42.3_{\pm 2.5}$ & $36.7_{\pm 2.1}$ & $34.3_{\pm 2.1}$ & $44.1_{\pm 2.1}$ \\
\tabledivider
RQGM & $\underline{50.3}_{\pm 1.5}$ & $\underline{42.7}_{\pm 2.1}$ & $\underline{61.7}_{\pm 2.1}$ & $\underline{43.7}_{\pm 2.5}$ & $\underline{37.3}_{\pm 1.5}$ & $\underline{35.0}_{\pm 2.0}$ & $\underline{45.1}_{\pm 1.7}$ \\
\rowcolor{vbdrow}\textbf{VbD (ours)} & $\mathbf{53.3}_{\pm 2.5}$ & $\mathbf{45.7}_{\pm 2.1}$ & $\mathbf{65.0}_{\pm 2.6}$ & $\mathbf{46.3}_{\pm 2.5}$ & $\mathbf{41.7}_{\pm 2.1}$ & $\mathbf{38.3}_{\pm 2.1}$ & $\mathbf{48.4}_{\pm 2.3}$ \\
\tabledivider
\rowcolor{black!6}External feedback$^{\dagger}$ & $62.0_{\pm 2.0}$ & $55.7_{\pm 1.2}$ & $69.7_{\pm 2.5}$ & $52.3_{\pm 2.1}$ & $50.3_{\pm 1.2}$ & $44.7_{\pm 1.2}$ & $55.8_{\pm 1.4}$ \\
\bottomrule
}
\setlength{\tabcolsep}{0pt}
\setbox0=\hbox{\begin{tabular}{l*{7}{r}}\qwencoloredrows\end{tabular}}
\setlength{\tabcolsep}{\dimexpr(\linewidth-\wd0)/16\relax}
\begin{tabular}{l*{7}{r}}
\qwencoloredrows
\end{tabular}
\tablenote{Overall pools outcomes across the six equally sized domains within each run. HR denotes human resources.}
\endgroup
\end{table*}

%% file: tables/qwen_next_tau3.tex
\begin{table}[t]
\caption{\textbf{Qwen3.8-Flash-Next results on $\tau^3$ Banking (\%; mean $\pm$ SD).} Values show the mean and sample SD across three runs; $\Delta$ is the gain over Base model in percentage points, computed before rounding.}
\label{tab:qwen-tau3}
\centering
\begingroup
\tablestylesetup
\small
\setlength{\tabcolsep}{0pt}
\newcommand{\qwentaurows}{%
\toprule
\textbf{Method} & \textbf{Pass rate (\%)} & $\Delta$ \\
\midrule
Base model & $43.0_{\pm 3.1}$ & $--$ \\
BoN & $49.8_{\pm 2.4}$ & $6.9$ \\
\tabledivider
HGM & $50.5_{\pm 1.8}$ & $7.6$ \\
RHO & $49.1_{\pm 2.4}$ & $6.2$ \\
RHI & $\underline{53.6}_{\pm 2.1}$ & $10.7$ \\
\tabledivider
RQGM & $51.9_{\pm 1.2}$ & $8.9$ \\
\rowcolor{vbdrow}\textbf{VbD (ours)} & $\mathbf{57.4}_{\pm 2.6}$ & $14.4$ \\
\tabledivider
\rowcolor{black!6}External feedback$^{\dagger}$ & $63.9_{\pm 2.1}$ & $21.0$ \\
\bottomrule
}
\setbox0=\hbox{\begin{tabular}{lrr}\qwentaurows\end{tabular}}
\setlength{\tabcolsep}{\dimexpr(\linewidth-\wd0)/6\relax}
\begin{tabular}{lrr}
\qwentaurows
\end{tabular}
\endgroup
\end{table}

%% file: tables/qwen_next_gdpval.tex
\begin{table*}[t]
\caption{\textbf{Qwen3.8-Flash-Next results on GDPval (rubric quality, \%; mean $\pm$ SD).} Scoring and notation follow Table~\ref{tab:gdpval-results}; aggregation is specified in Appendix~\ref{app:seed-reporting}.}
\label{tab:qwen-gdpval}
\centering
\begingroup
\tablestylesetup
\footnotesize
\newcommand{\qwencoloredrows}{%
\toprule
\textbf{Method} & \textbf{Prof} & \textbf{Govt} & \textbf{Manuf} & \textbf{RealEst} & \textbf{Fin} & \textbf{Info} & \textbf{Whsle} & \textbf{Health} & \textbf{Retail} & \textbf{Overall} \\
\midrule
Base model & $67.1_{\pm 2.6}$ & $73.1_{\pm 2.0}$ & $77.5_{\pm 0.2}$ & $73.6_{\pm 1.2}$ & $71.6_{\pm 2.6}$ & $75.6_{\pm 2.6}$ & $72.4_{\pm 1.2}$ & $71.1_{\pm 1.4}$ & $73.2_{\pm 2.4}$ & $72.3_{\pm 1.7}$ \\
BoN & $78.1_{\pm 2.7}$ & $79.7_{\pm 2.7}$ & $82.1_{\pm 1.2}$ & $80.7_{\pm 1.4}$ & $80.8_{\pm 2.7}$ & $82.8_{\pm 2.4}$ & $80.1_{\pm 0.4}$ & $79.2_{\pm 2.6}$ & $79.1_{\pm 2.2}$ & $80.0_{\pm 1.9}$ \\
\tabledivider
HGM & $78.8_{\pm 2.8}$ & $81.3_{\pm 2.4}$ & $\underline{85.6}_{\pm 2.7}$ & $82.5_{\pm 2.8}$ & $81.5_{\pm 2.7}$ & $\underline{83.7}_{\pm 2.5}$ & $79.9_{\pm 2.5}$ & $80.3_{\pm 2.8}$ & $81.7_{\pm 0.8}$ & $81.3_{\pm 2.4}$ \\
RHO & $80.3_{\pm 1.2}$ & $81.8_{\pm 2.8}$ & $83.3_{\pm 2.7}$ & $81.6_{\pm 2.8}$ & $80.7_{\pm 0.4}$ & $82.1_{\pm 0.8}$ & $80.9_{\pm 2.2}$ & $79.7_{\pm 1.6}$ & $81.3_{\pm 2.7}$ & $81.2_{\pm 1.7}$ \\
RHI & $81.2_{\pm 2.5}$ & $\underline{83.9}_{\pm 2.6}$ & $84.2_{\pm 2.1}$ & $83.5_{\pm 1.9}$ & $82.6_{\pm 2.7}$ & $82.7_{\pm 2.0}$ & $81.6_{\pm 2.2}$ & $80.5_{\pm 0.6}$ & $83.3_{\pm 2.5}$ & $82.4_{\pm 2.0}$ \\
\tabledivider
RQGM & $\underline{82.8}_{\pm 1.0}$ & $81.3_{\pm 2.1}$ & $84.1_{\pm 2.8}$ & $\underline{83.7}_{\pm 1.0}$ & $\underline{83.1}_{\pm 2.8}$ & $82.3_{\pm 1.2}$ & $\underline{83.1}_{\pm 2.8}$ & $\underline{83.7}_{\pm 1.4}$ & $\underline{84.9}_{\pm 0.9}$ & $\underline{83.1}_{\pm 1.6}$ \\
\rowcolor{vbdrow}\textbf{VbD} & $\mathbf{84.8}_{\pm 1.5}$ & $\mathbf{84.1}_{\pm 2.9}$ & $\mathbf{86.6}_{\pm 2.7}$ & $\mathbf{84.0}_{\pm 1.5}$ & $\mathbf{84.5}_{\pm 2.9}$ & $\mathbf{84.3}_{\pm 1.3}$ & $\mathbf{84.2}_{\pm 2.8}$ & $\mathbf{84.3}_{\pm 0.6}$ & $\mathbf{85.7}_{\pm 1.9}$ & $\mathbf{84.7}_{\pm 2.0}$ \\
\tabledivider
\rowcolor{black!6}External feedback$^{\dagger}$ & $85.8_{\pm 2.6}$ & $86.4_{\pm 1.6}$ & $88.6_{\pm 2.1}$ & $86.5_{\pm 1.6}$ & $86.1_{\pm 2.5}$ & $87.1_{\pm 1.6}$ & $85.7_{\pm 1.2}$ & $86.7_{\pm 0.7}$ & $87.5_{\pm 2.8}$ & $86.6_{\pm 1.7}$ \\
\bottomrule
}
\setlength{\tabcolsep}{0pt}
\setbox0=\hbox{\begin{tabular}{l*{10}{r}}\qwencoloredrows\end{tabular}}
\setlength{\tabcolsep}{\dimexpr(\linewidth-\wd0)/22\relax}
\begin{tabular}{l*{10}{r}}
\qwencoloredrows
\end{tabular}
\tablenote{Sector labels: Prof = professional services; Govt = government; Manuf = manufacturing; RealEst = real estate; Fin = finance; Info = information; Whsle = wholesale; Health = healthcare; Retail = retail.}
\endgroup
\end{table*}

%% file: sections/appendix_partial_credit.tex
\subsection{Partial Task Completion}
\label{app:partial-credit}
Pass rate credits a task only when all scored checks pass. AutomationBench also provides partial credit, the fraction of scored assertions satisfied for each task~\citep{shepard2026automationbench}. Averaging these fractions reveals progress that binary task completion can conceal.

\begin{figure}[t]
\centering
\includegraphics[width=\linewidth]{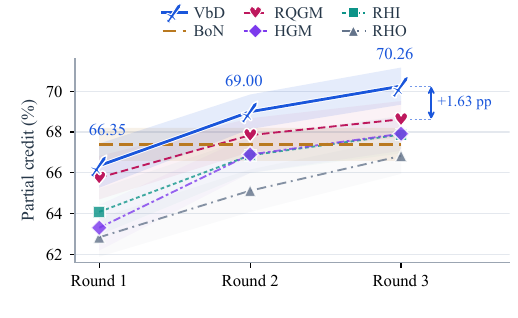}
\caption{\textbf{Partial credit on AutomationBench.} Scores average the fraction of satisfied assertions per task with DeepSeek-V4-Flash-0731. BoN is shown as a horizontal baseline. Shading shows $\pm$ SD.}
\label{fig:partial-credit}
\end{figure}

Figure~\ref{fig:partial-credit} shows VbD increasing from 66.35\% to 70.26\%, a gain of 3.91 percentage points. RHO increases from 62.83\% to 66.83\%. VbD exceeds BoN from Round~2 onward and finishes 1.63 percentage points above the strongest baseline, RQGM. This complements the results on pass rate: later outputs satisfy a larger fraction of \mbox{scored requirements}.

%% file: sections/appendix_oracle.tex
\subsection{Evolution with External Feedback}
\label{app:oracle-evolution}

The external feedback reference uses HGM to evolve the Actor under fixed model weights, with benchmark scores guiding updates and output selection. AutomationBench and $\tau^3$ Banking supply outcome checks; GDPval supplies local rubric scores. VbD and other baselines use these scores only offline.

\label{app:oracle-comparison}
Tables~\ref{tab:main-results}--\ref{tab:gdpval-results} compare the same task sets. The reference generates its own candidates, whereas an oracle-selection bound requires a shared candidate pool~\citep{brown2024monkeys}.

%% file: tables/gating_counts.tex
\begin{table}[t]
\caption{\textbf{Actor gating counts.} With DeepSeek-V4-Flash-0731 on AutomationBench, proposed and accepted totals cover all updates; TP, FP, and TN cover valid, non-tied comparisons.}
\label{tab:gating-counts}
\centering
\begingroup
\tablestylesetup
\setlength{\tabcolsep}{3pt}
\renewcommand{\arraystretch}{1.05}
\begin{tabularx}{\linewidth}{@{}lrr*{3}{R}@{}}
\toprule
Method & Proposed & Accepted & TP & FP & TN \\
\midrule
HGM & 1,200 & 227 & 158 & 56 & 108 \\
RHO & 1,432 & 698 & 302 & 121 & 89 \\
RHI & 1,723 & 1,412 & 198 & 240 & 97 \\
RQGM & 1,200 & 231 & 164 & 49 & 121 \\
\rowcolor{vbdrow}\textbf{VbD} & 1,104 & 207 & 156 & 45 & 374 \\
\bottomrule
\end{tabularx}
\endgroup
\end{table}

%% file: sections/appendix_learned.tex
\Needspace{16\baselineskip}
\section{Behavioral Dynamics of Evolution}
\label{app:learned-harnesses}

\begin{figure*}[!t]
\centering
\includegraphics[width=\textwidth]{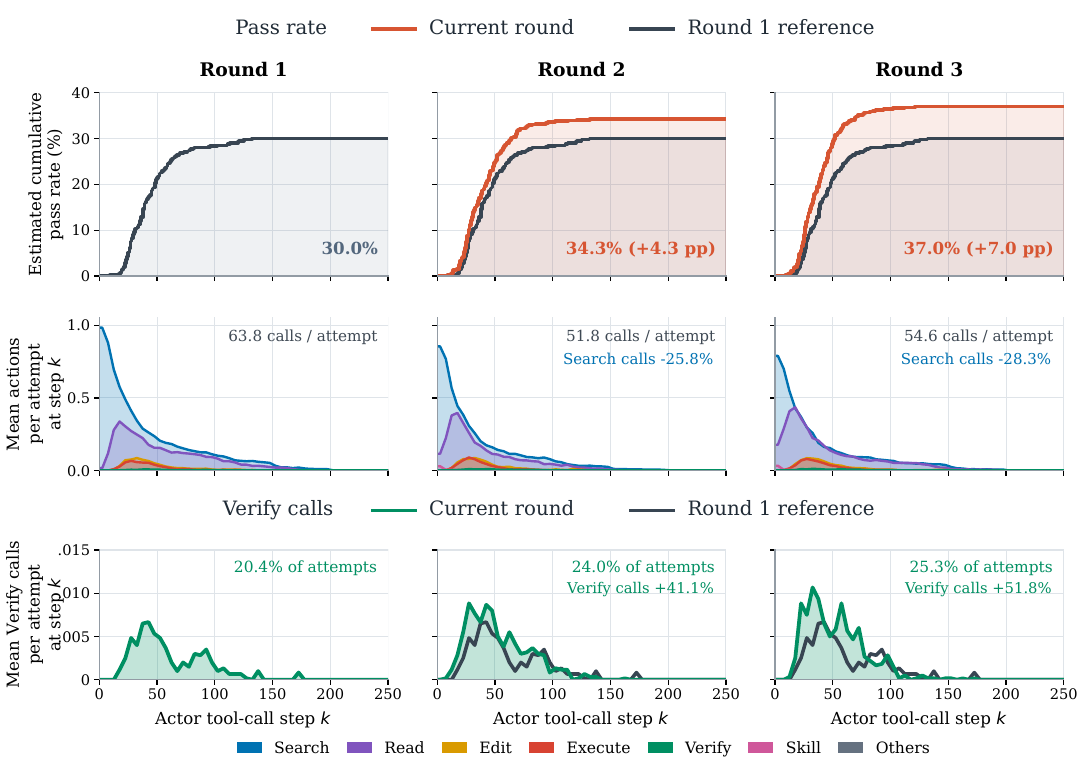}
\caption{\textbf{Selected outcomes and Actor behavior in Rounds~1--3.} For DeepSeek-V4-Flash-0731 on AutomationBench, the top row estimates passing selected outputs by tool-call count, normalized to round totals. The middle row shows actions per attempt in five-call bins; the bottom row magnifies Verify frequency. Dark gray curves show the Round~1 reference. Appendix~\ref{app:actor-behavior} defines the measurements.}
\label{fig:behavior-details}
\label{fig:behavior-overview}
\end{figure*}

Figures~\ref{fig:behavior-details}--\ref{fig:behavior-domains} and Table~\ref{tab:actor-rounds} analyze DeepSeek-V4-Flash-0731; Figures~\ref{fig:baseline-behavior}--\ref{fig:task-space-comparison} use Qwen3.8-Flash-Next. Skill loading records explicit use of evolved guidance, while Verify calls capture checks of action effects. These measurements connect the case studies to aggregate changes in Actor behavior.

\input{tables/actor_behavior_summary}

\subsection{Skill Use and Verification Behavior}
\label{app:behavior-change}

For findings to improve execution, later Actors must use the guidance produced by harness evolution. The share of attempts loading skills rises from 0.0\% in Round~1 to 23.1\% in Round~3 (Table~\ref{tab:actor-rounds}). This use of explicit guidance complements the cases in which a discovered operation or missed requirement becomes a reusable Actor skill.

\begin{figure*}[!t]
\centering
\includegraphics[width=\textwidth]{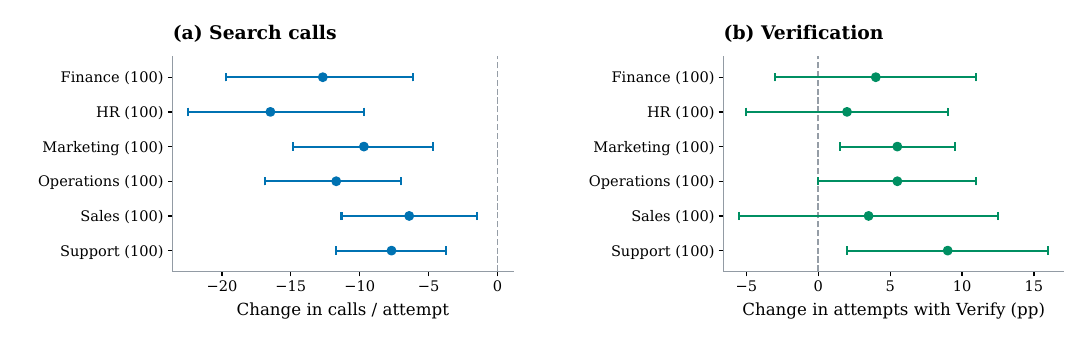}
\caption{\textbf{Behavioral changes by domain.} For DeepSeek-V4-Flash-0731 on AutomationBench, panels show changes from Round~1 to Round~3 in Search calls per attempt (left) and attempts with verification (right, pp). Whiskers show 95\% bootstrap intervals over tasks.}
\label{fig:behavior-domains}
\end{figure*}

Later rounds use fewer Search calls and more Verify calls. Attempts containing state verification rise from 20.4\% to 25.3\% between Rounds~1 and~3, while total calls per attempt fall from 63.8 to 54.6 (Table~\ref{tab:actor-rounds}). Thus, more attempts include state checks even as the average execution uses fewer calls. Search decreases in every domain, while verification increases vary across domains (Figure~\ref{fig:behavior-domains}).

\begin{figure*}[tp]
\centering
\includegraphics[width=\textwidth]{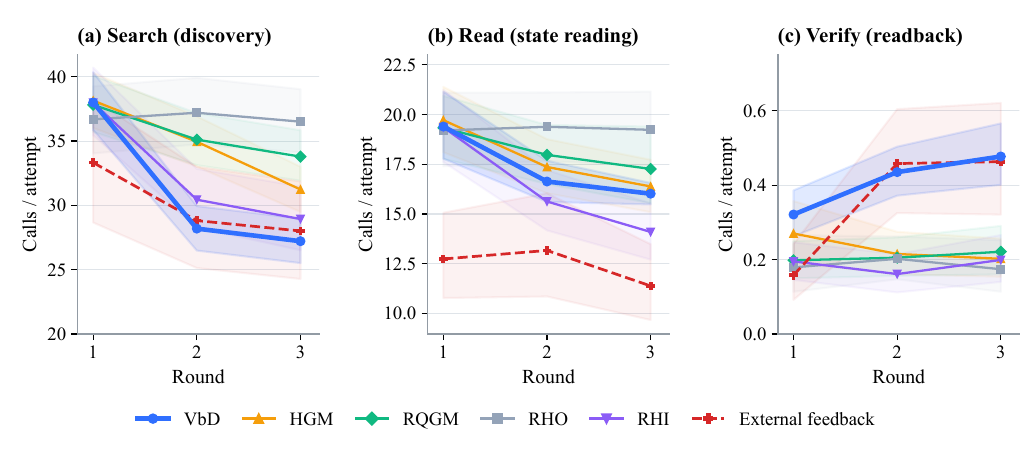}
\caption{\textbf{Actor behavior across rounds on AutomationBench.} Curves show Search, Read, and Verify calls per attempt with Qwen3.8-Flash-Next for VbD, four evolution baselines, and the external feedback reference (dashed). Shading shows 95\% bootstrap intervals over tasks. Appendix~\ref{app:actor-behavior} details the measurements.}
\label{fig:baseline-behavior}
\end{figure*}

The cross-method comparison with Qwen3.8-Flash-Next places VbD's increased checking in context (Figure~\ref{fig:baseline-behavior}). HGM and RHI also reduce Search and Read, but their Verify frequency remains flat or decreases.
\newpage
RQGM shows a smaller change in Verify, while RHO remains relatively stable across all three categories. VbD combines reduced Search with increased Verify, a direction also seen in the external feedback reference. This pattern is consistent with VbD's design: verification findings guide both output assessment and checks performed during subsequent execution. Appendix~\ref{app:actor-behavior} details the measurements.

These observations connect the Actor side of VbD to the Evaluator results. On fixed candidates, the evolved Evaluator selects better outputs (Appendix~\ref{app:evaluator-protocol}); in later executions, Actors use evolved skills and check state more often. The cases trace how shared findings link these changes, while the ablations support both acquiring evidence and retaining improved checks (Section~\ref{sec:ablations}). Together, these results support VbD's central design: improve execution and the verification procedure that supplies its internal signal within the same task.

\newpage
\null\vfill\newpage
\null\vfill\newpage
\subsection{Task Outcome Visualization}
\label{app:task-projection}

\begin{figure}[!htbp]
  \centering
  \includegraphics[width=\linewidth]{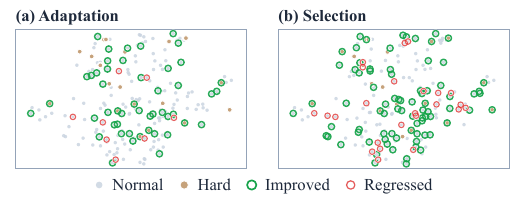}
  \caption{\textbf{Task outcomes in t-SNE space.} Using Qwen3.8-Flash-Next on 200 AutomationBench tasks, panels compare (a) initialization with final selection and (b) VbD with self-preference. Orange (Hard) marks VbD successes missed by the reference with external feedback in this comparison of individual tasks; gray (Normal) provides context. Green/red rings indicate label gains/losses across the same tasks.}
  \label{fig:task-space-comparison}
\end{figure}

The task projection displays Qwen3.8-Flash-Next outcomes on individual AutomationBench tasks. Figure~\ref{fig:task-space-comparison} projects TF-IDF task descriptions using 50-dimensional TruncatedSVD, normalization, and t-SNE with PCA initialization and perplexity 35. Both panels share 200 tasks and coordinates; outcome labels determine display inclusion and annotations, while the projection uses task text. Orange highlights VbD successes missed by all six candidates from the external feedback reference. ``Hard'' and ``Normal'' are display labels for these highlighted tasks and their context.

\input{sections/appendix_behavior}

%% file: tables/actor_behavior_summary.tex
\begin{table}[!ht]
\caption{\textbf{AutomationBench Actor behavior.} The table reports calls per attempt and rates of verification and skill loading for DeepSeek-V4-Flash-0731, following Appendix~\ref{app:actor-behavior}.}
\label{tab:actor-rounds}
\centering\tablestylesetup
\setlength{\tabcolsep}{2pt}
\begin{tabularx}{\linewidth}{@{}P{.50\linewidth}*{3}{R}@{}}
\toprule
\textbf{Metric} & \textbf{Round~1} & \textbf{Round~2} & \textbf{Round~3} \\
\midrule
Tool calls per attempt & 63.8 & 51.8 & 54.6 \\
Attempts with Verify (\%) & 20.4 & 24.0 & 25.3 \\
Attempts loading skills (\%) & 0.0 & 22.2 & 23.1 \\
\bottomrule
\end{tabularx}
\end{table}

%% file: sections/appendix_behavior.tex
\subsection{Behavioral Measurement Protocol}
\label{app:actor-behavior}

For each method and backbone, we analyze both Actor executions per round on all 600 AutomationBench tasks across three runs, including failed attempts and rejected proposals. We average counts and attempt-level rates within tasks, then across tasks.

Table~\ref{tab:actor-taxonomy} extends RHO's action categories~\citep{pan2026rho} with explicit skill loading and other calls. Verify takes precedence over Read and Search.

\begin{table}[htbp]
\caption{\textbf{Actor action categories.} AutomationBench tool calls; Verify marks state checks after actions.}
\label{tab:actor-taxonomy}
\centering\tablestylesetup
\begin{tabularx}{\linewidth}{lL}
\toprule
Action & Recorded operation \\
\midrule
Search & Discover APIs or search for relevant records. \\
Read & Retrieve a specified resource or state. \\
Edit & Create, update, or delete records or artifacts. \\
Execute & Send, publish, submit, or transform data. \\
Verify & Read the state after an earlier action. \\
Skill & Explicitly load a skill. \\
Others & Tool names outside the standard taxonomy. \\
\bottomrule
\end{tabularx}
\end{table}

Verify counts later read-backs of the same object after an action without an explicit error; source consultation and prerequisite checks remain Search or Read. All behavioral analyses use the same object-matching rule. Verification and skill-loading rates count attempts with at least one corresponding call.

\paragraph{Within-round trajectories.}
In Figure~\ref{fig:behavior-details}, step $k$ counts tool calls within an attempt. The top row estimates the fraction of tasks with a passing selected output generated within $k$ calls, normalized to each round's pass rate. These curves follow the selected output's original execution, excluding discarded attempts and controller work.

The middle row reports mean calls per attempt per position in five-call bins; ended attempts contribute zero. The bottom row magnifies Verify on the same basis. Count changes use relative percentages; share and attempt-rate changes use percentage points.

Behavioral intervals resample tasks, retaining each task's attempts and rounds together. Initial-to-final changes use within-task differences, and domain intervals resample within domains. Paired outcome comparisons retain all 600 tasks, with unresolved grades receiving no pass credit.

%% file: sections/appendix_harness_guide.tex
\Needspace{12\baselineskip}
\section{Harness Specification}
\label{app:ab-prompts}

The harnesses express task requirements as procedures for execution and verification. Role instructions define how each procedure operates; recorded updates and executed checks show how an observed gap becomes guidance that the agent can use.

\subsection{Role Policies and Verification}
Listings~\ref{lst:actor-prompt}--\ref{lst:actor-applicability} give the Actor instructions, and Listing~\ref{lst:evaluator-prompt} gives the Evaluator instructions. These accompany the task, working rubric, available skills and tools, and assigned checks. Tool availability follows the evaluated condition.

The Evaluator inspects unresolved requirements using the instructions and tools in Listings~\ref{lst:evaluator-runtime}--\ref{lst:verification-tools}. Returned observations guide further inquiry until completion or budget exhaustion.

Skills check delivered effects and rule applicability, including competing interpretations (Listings~\ref{lst:skill-0}--\ref{lst:skill-4}). Both roles' rubrics specify a criterion, a verification plan naming the needed observations, a falsifier identifying contrary evidence, and an unknown rule for unresolved evidence (Listings~\ref{lst:rubric-1}--\ref{lst:rubric-6}).

\subsection{Updates, Selection, and Recorded Use}
The joint editor proposes changes and expected observations (Listing~\ref{lst:joint-editor}). Listings~\ref{lst:reviewer}--\ref{lst:artifact-preference} specify Evaluator review and output preference; Appendix~\ref{app:execution-configuration} defines trials and retention.

\label{app:resource-checks}
In Marketing015, a successful lookup becomes an Actor retrieval instruction and an Evaluator availability check (Listings~\ref{lst:resource-revision}--\ref{lst:resource-execution}).
\label{app:delivered-content-checks}
Support091 instead pairs an adopted content criterion with message reads that change selection between existing outputs (Listings~\ref{lst:content-rubric}--\ref{lst:content-execution}). Sales063 updates both roles around the reply restriction: an Actor skill checks reply scope and an Evaluator rubric assesses the same requirement (Listings~\ref{lst:actor-skill}--\ref{lst:paired-working-checks}). Listing~\ref{lst:joint-hypothesis} records the rationale and expected observations for that proposal.

\label{app:environment-checks}
Two further records show why verification must remain tied to a candidate's own environment. Sales052 checks whether a policy was available and reads the delivered message before judging its omission (Listing~\ref{lst:ab-policy-content}). Finance097 establishes that a newer status update was available to the candidate and checks the delivered journal; the observation precedes the skill adopted in the following round (Listing~\ref{lst:ab-recent-updates}). The listings preserve the requests, returned evidence, and retention decisions. Their Round~3 trials record no explicit skill loading; the observations describe the full Evaluator procedure.

\subsection{Deliverables and Executable Checks}
\label{app:deliverable-harnesses}
For GDPval, the Actor creates and reopens deliverables; the Evaluator inspects their contents and native structure (Listings~\ref{lst:gdp-actor}--\ref{lst:gdp-evaluator}).

\label{app:numerical-skills}
Listings~\ref{lst:budget-skill}--\ref{lst:grant-skill} specify the verification skills adopted for numerical requirements, with an executed date-and-threshold check recorded in Listing~\ref{lst:grant-program}.
\label{app:python-checks}
The remaining examples reproduce executed programs for source coverage, native workbook structure, and numerical consistency (Listings~\ref{lst:award-population-check}--\ref{lst:capacity-balance-check}). Award reconciliation compares the complete qualifying ID set with the delivered workbook. Pivot inspection tests for native workbook components that displayed totals cannot establish. Capacity verification recomputes a recurrence over the entered demand and capacity. Each program is paired with its returned observation, so the reader can see both the checked property and the inputs on which the conclusion depends.

%% file: sections/appendix_prompts.tex
\begin{harnessbox}{Actor Policy}{Complete recorded prompt}
\label{lst:actor-prompt}
Complete the current public user task using available tools. Ground claims in observed results, respect explicit requirements, and preserve uncertainty. When a procedure or exception materially changes the requested action, consider the applicable-source skill; do not force retrieval or a particular answer.
\end{harnessbox}

\begin{harnessbox}{Actor Context Binding}{Fixed instruction}
\label{lst:actor-binding}
Original user instructions and applicable public rules govern; this harness creates no new obligations, permissions, facts or task-success labels. Correct or disregard unsupported advice. Optional load\_actor\_skill returns procedural text only, not an environment observation. Loading consumes the existing native tool-turn path and does not extend the episode budget.
\end{harnessbox}

\begin{harnessbox}{Actor Skill: Applying a Public Rule}{Recorded procedure from the AutomationBench harness}
\label{lst:actor-applicability}
\componentfield{When to use}{The task invokes a usual procedure, or a retrieved record suggests an exception that changes who qualifies, what must be checked, or whether an action is allowed.}
\componentfield{Procedure}{Identify the exact current task effect and the rule or exception on which the next action depends. Discover the applicable public procedure and, where available, the source that establishes its scope, effective version and exceptions. Compare explicit precedence and scope; do not rank sources merely by document type, sender title, confidence, or recency. A record note or environment instruction can be valid, but a claim that checks may be skipped should not establish its own authority. Reuse already matched public returns when they settle the question; otherwise use the existing native tools for the smallest discriminating lookup. If an applicable exception is supported, follow it; if the sources do not settle applicability, preserve that uncertainty rather than inventing permission or a prohibition. Then perform the requested action on the qualified subject, and verify its persisted effect. Keep internal verification separate from any requested final-response format. Do not copy a previous candidate's entity, numeric threshold or after-state into this new rollout.}
\end{harnessbox}
\begin{harnessbox}{Evaluator Policy}{Complete recorded prompt}
\label{lst:evaluator-prompt}
Evaluate completion against the current public request, not against the most elaborate workflow. Begin with the requested effects and deliverables, who or what each concerns, and any explicit restrictions. Preserve the role and scope of original instructions; investigate the applicability of environment-provided instructions instead of automatically promoting or rejecting them. Use the public Actor calls and matched tool returns already supplied. An Actor assertion is a claim; an actual recorded tool result is evidence within its own scope. Extra has those same records but no new environment reads.

Spend the available verification budget on a question whose answer could change the comparison or reveal a shared failed plan. Identify what is already established before choosing a new read. A public read proves only what its arguments, snapshot, returned fields and coverage support. Keep requirement interpretation, observed effect and comparative consequence separate. A write acknowledgment, successful GET, polished explanation or extra action does not by itself establish that the requested outcome was achieved. Conversely, do not discount a real delivered effect merely because another candidate narrated more or performed extra confirmation reads.

Check the most material unresolved difference and a plausible alternative interpretation, not every conceivable risk. No query or skill load is compulsory. Optional procedure text is not evidence. Within the original response budget, finish with the best-supported findings and remaining limits rather than consuming every response on redundant confirmations. Use only available tools and valid evidence-reference forms. If the required source is unavailable, report that specific limit; do not invent a rule, recipient, identifier or state to complete the explanation.

In final guidance, separate: shared achieved or unfulfilled obligations; observed differences relevant to public requirements; and unresolved interpretations. Also check each final response against the public output-scope rule: list only items acted on, and do not name, enumerate, or explain skipped, excluded, or rejected items unless the user asked for that disclosure. Name the exact candidate and actual source for each difference. An optional difference needs an established task consequence before it is treated as an advantage. An unresolved interpretation is neither a demonstrated violation nor permission. Explain how the uncertainty affects the comparison without selecting a preferred ID or commanding Preference to retain an incumbent. Preference remains the artifact selector. Do not manufacture a completion advantage when both candidates have the same unsupported plan; also do not turn factual equivalence into a universal prohibition on alternative \mbox{valid implementations.}
\end{harnessbox}
\begin{harnessbox}{Evaluator Interaction Instructions}{Fixed instruction excerpt}
\label{lst:evaluator-runtime}
Evaluate the supplied public task and candidate, not an official or hidden grading standard. You can actually call the provided tools; their results will be returned in this conversation. Examine consequential uncertainty. A test may be wrong and one implementation need not be mandatory. A successful read is not proof of task success. Self-reviews are model judgments, not new facts. Do not infer that the original Agent made your later queries. Do not choose an answer \mbox{or change a playbook.}
\end{harnessbox}

\begin{harnessbox}{AutomationBench Verification Tools}{Available operations}
\label{lst:verification-tools}
\componentfield{Discover an operation}{Find an available application operation and its required inputs. Bind the request to entities identified in the task or observed records.}
\componentfield{Inspect an effect}{Read the designated candidate's saved state. When a pre-execution snapshot is available, compare it with the resulting state to distinguish existing records from the candidate's actions.}
\componentfield{Load a procedure}{Read a relevant skill and apply its checks to the available evidence. Loading uses the existing interaction budget; the Actor can also load execution skills.}
\componentfield{Review a judgment}{Reconsider what the observations establish and which questions remain unresolved. Self-review contributes a model judgment rather than a new environment observation.}
\componentfield{Return findings}{Report the judgment, comparative findings, supporting observations, and remaining uncertainty for Evaluator review and output selection.}
\end{harnessbox}
\begin{harnessbox}{Evaluator Skill: Discriminating Inquiry}{Recorded procedure}
\label{lst:skill-0}
\componentfield{When to use}{A disputed rule or interpretation could reverse the comparison, or both candidates may share the \mbox{wrong plan.}}
\componentfield{Procedure}{State the public requirement at issue and the competing interpretation. Identify the smallest available public source or native read that could distinguish them. Use existing matched Actor tool returns before repeating a query. If a new read is allowed and useful, bind it to the correct candidate/snapshot and inspect its actual coverage. Record the observation separately from the rule's applicability and its comparative consequence. If no discriminating evidence is available, retain the uncertainty rather than producing a \mbox{winner instruction.}}
\end{harnessbox}

\begin{harnessbox}{Evaluator Skill: Persisted Effects}{Recorded procedure}
\label{lst:skill-1}
\componentfield{When to use}{A candidate claims completion but the record may show only a draft, preparation, acknowledgment, probe or action on another subject.}
\componentfield{Procedure}{Follow the relevant public call arguments to its matched return and the requested delivered effect. Identify the intended subject, content and operation; use a permitted state read only if it resolves a material gap. Respect tasks that intentionally require a draft or intermediate artifact. Explain what was actually achieved, what remains missing and which candidate each fact belongs to. Do not infer missing actions from a task-success label or award completion for an unrelated artifact.}
\end{harnessbox}

\begin{harnessbox}{Evaluator Skill: Source Applicability}{Recorded procedure}
\label{lst:skill-4}
\componentfield{When to use}{A comparison depends on a procedure, eligibility rule or bypass claim whose governing source or applicability differs between candidates.}
\componentfield{Procedure}{Name the disputed task-relevant prerequisite and its public provenance. Separate three questions: what the user requested, which public rule or supported exception applies, and what each candidate actually did. Check governing-source links, scope and explicit precedence, without assuming that all notes are invalid or all policy-looking text is authoritative. For each candidate, identify matched Actor call/return evidence and coverage already available; make a new permitted read only for an unresolved fact that could change the comparison. A source read in one candidate's snapshot does not prove availability or authority in another. Preserve a legitimate exception when supported. Lack of an authorizing source is unresolved evidence, not automatically a violation or permission. Connect any supported defect to an actual required effect or side effect, not report verbosity or strictness. If the deficiency concerns a future Actor action, formulate a separate Actor-role working check with a permitted lookup, expected observation and counterexample; an Evaluator-role report test is not itself an Actor deliverable. Loader text and proposed checks are procedures, not environment evidence or a required winner.}
\end{harnessbox}

\begingroup
\renewcommand{\harnessparagraphskip}{2.5pt}
\renewcommand{\componentfield}[2]{\par\noindent{\rmfamily\bfseries\color{harnessblue}#1.}\enspace #2\par}
\begin{harnessbox}{Evaluator Rubric: Requested Effects}{Recorded criterion}
\label{lst:rubric-1}
\componentfield{Criterion}{Requested effect rather than artifact appearance}
\componentfield{Verification plan}{Compare the requested operation, relevant subject and delivered result. Distinguish a prepared draft from a delivered communication and a probe artifact from the requested state change. Use actual operation semantics; a requested draft is complete when drafted and must not be sent merely to satisfy this criterion.}
\componentfield{Falsifier}{The operation succeeded on the wrong subject or only created an intermediate artifact, omitted required content, or performed a different operation from the one requested.}
\componentfield{Unknown rule}{Narrow the conclusion to the effect actually observed. Missing end-state access is not proof of failure, and a successful return is not proof of every requested consequence.}
\end{harnessbox}

\begin{harnessbox}{Evaluator Rubric: Candidate Attribution}{Recorded criterion}
\label{lst:rubric-2}
\componentfield{Criterion}{Candidate and temporal attribution}
\componentfield{Verification plan}{Bind calls, returns, final responses and observations to their exact candidate and before/after snapshot. Support statements about both candidates with evidence about each. Re-establish old observations for a new rollout rather than carrying them forward as current facts.}
\componentfield{Falsifier}{A behavior is attributed to the other candidate, a before-state is treated as the final result, or an earlier candidate's count is contradicted using a different later candidate.}
\componentfield{Unknown rule}{Keep the supported single-candidate claim when coverage is asymmetric. An unavailable record for the other candidate remains unknown, not copied from the first.}
\end{harnessbox}

\begingroup

\begin{harnessbox}{Evaluator Rubric: Rule Applicability}{Recorded criterion}
\label{lst:rubric-6}
\componentfield{Criterion}{Applicable procedure, supported exceptions, and the requested persisted effect}
\componentfield{Verification plan}{Trace the task obligation to applicable public sources, check any claimed exception at its proper scope, and compare each candidate's matched action/results against that obligation. Reuse sufficient recorded evidence before \mbox{requesting another read.}}
\componentfield{Falsifier}{Public evidence shows the candidate used an inapplicable rule or unsupported exception and thereby failed a requested effect; conversely a supported exception or sufficient existing evidence can refute a proposed defect.}
\componentfield{Unknown rule}{An unsettled authority or applicability question remains unknown. Do not invent a prohibition, permission, strict comparative advantage or automatic incumbent retention.}
\end{harnessbox}

\endgroup

\endgroup
\begin{harnessbox}{Joint Editor}{Fixed instruction excerpt for fresh trials}
\label{lst:joint-editor}
Improve this same-task Actor/Evaluator system around ONE testable improvement hypothesis, not one field. You may make a substantial coordinated change to ANY SUBSET of editable Actor and Evaluator prompt, rubric and skills modules. Choose the amplitude required by the problem; neither minimal edits nor rewriting every module is intrinsically better. An empty module\_changes list is also valid. Different rounds may explore different ideas. Declare one entry per changed/reused module and its source\_module\_ref from actual accepted or archived pointers. A rejected proposal may still be a useful exploration parent; read its public counterevidence before reusing it. Set replacement=null to reuse that exact module value, or provide a revised value for that module. The executor preserves only modules you did not declare; it does not restrict your bundle to one module or one role. A composition is only a proposed trial, not an automatically accepted improvement. Explain why all bundle edits jointly address the same problem, the observable predicted effect, and a concrete counterexample. The schema does not certify semantic coherence; the public trial must test the bundle.

External feedback from official task scoring is NOT available or editable. Verification procedures, rubrics, and instructions can evolve to improve the internal signal under unchanged original public objectives. Historical preference outcomes belong to their original candidate IDs and criterion version, not a comparable current success score. More strict judgments, report length, query counts or agreement with yourself are \mbox{not improvement.}

Separately retain zero to two working\_checks, at most one per editable role, even when the persistent edit is empty. Every role with a selected module requires a check. Each check must identify an actual public obligation/uncertainty, a permitted discriminating test, an expected observation, a counterexample and unknown handling. Cite public origin refs, not the module catalog as task authority. Respect requested output and side-effect limits; a question about an internal uncertainty need not add narration to the final answer. Checks last one next round unless reconsidered from fresh evidence. No forced GET.

Both next Actors see the same checks; only the proposed-policy Actor uses the composed Actor module. Parent and proposed Evaluators test the same fresh pair and questions under the existing budget. Existing qualification and Actor promotion remain unchanged. A skill must be loaded and its procedure followed to establish an observation; loading is not environment evidence. No external memory, new tools, reference labels or task answers.
\end{harnessbox}
\begin{harnessbox}{Evaluator Reviewer}{Fixed instruction excerpt}
\label{lst:reviewer}
Compare two evaluation reports on exactly the same current public candidate pair. Choose the report better supported by the public task, branch-scoped Actor messages and actual observations. This selects an evaluation procedure, NEVER an Actor artifact or task-success label. Distinguish a true record from an applicable instruction and from a justified comparative preference. Do not discard instructions merely because they arrive through the task environment. Earlier-candidate facts cannot establish current-candidate facts. More queries or confident prose is not better. Keep unresolved applicability unresolved without inventing either permissions or prohibitions. Each report has its own execution-reference namespace; do not mix them. If neither has a supported comparative advantage, choose undecided. No official score or hidden answer is available. Return exactly \{"choice":"first"|\allowbreak"second"|\allowbreak"undecided",\allowbreak"reason":string\}.
\end{harnessbox}

\begin{harnessbox}{Artifact Preference}{Fixed AutomationBench instruction excerpt}
\label{lst:artifact-preference}
Compare these two actual candidate artifacts against the original public task. Return relation=strict only when public evidence supports a material comparative advantage under the task requirements. Set preferred\_candidate\_id and representative\_candidate\_id to that same candidate. Equivalent means no supported material difference; undetermined means the relevant comparison is unresolved. For either non-strict relation preferred\_candidate\_id must be null, but choose a representative\_candidate\_id to submit. A representative is NOT evidence of policy improvement. Do not fabricate a strict advantage merely because one artifact is needed. Extra actions, confident wording or larger margins need an actual task consequence. Original public instructions retain authority. Rubrics, evaluator reports and procedure reviews are fallible advice, not new requirements or task-success labels. Treat actual Actor tool returns as scoped evidence; distinguish them from Actor assertions. Unresolved applicability is neither established violation nor permission. Preserve candidate and snapshot attribution and contrary evidence. No external feedback from official scoring is available. This is an artifact comparison, not evaluator-procedure qualification. Use the candidate IDs as presented and give the public reason for the relation.
\end{harnessbox}
\begingroup
\renewcommand{\componentfield}[2]{\par\noindent{\rmfamily\bfseries\color{harnessblue}#1.}\enspace #2\par}

\begin{harnessbox}{Resource Availability: Actor and Evaluator Updates}{Recorded Actor prompt excerpt and complete Evaluator skill; adopted in Round~2}
\label{lst:resource-revision}
\componentfield{Actor instruction}{When a task-referenced resource (for example a spreadsheet ID implied by the task or emails) is reported not found by one endpoint, check the API catalog for another endpoint on the same resource, such as a values.get or batchGet variant, and try that same ID before guessing alternate IDs or probing unrelated services. A not-found error from one endpoint variant is not evidence the resource is absent; a successful read from another variant is evidence of reachability, not of any action.}
\componentfield{Evaluator skill: when to use}{Both candidates failed to locate a task-referenced resource, or a candidate claims the resource is unavailable after only one endpoint variant returned not found.}
\componentfield{Evaluator skill: procedure}{Identify the resource ID the task or emails imply (e.g., media\_tracker). Check whether the API catalog exposes another endpoint for the same resource, such as sheets.spreadsheets.values.get or batchGet. If so, read the same ID with a plausible range and record the actual return. Use the result only to establish reachability and to explain a shared failed plan; do not treat the evaluator's successful read as evidence the candidate performed the action. If the alternate read also fails, keep the resource availability unresolved.}
\end{harnessbox}

\begin{harnessbox}{Press Release Distribution: Executed Verification}{Condensed execution record}
\label{lst:resource-execution}
\componentfield{Round~1: test the limitation}{The Actors receive ``Spreadsheet with id 'media\_tracker' not found.'' For each candidate's saved state, the Evaluator reads \texttt{media\_tracker} through \texttt{values/Sheet1!A1:Z100}. Both returns contain outlet rows, including TechWire Daily (Tier 1, DA 92) and CloudBeat (Tier 1, DA 88, with a litigation note). Mailbox reads and the Actor traces establish that neither attempt completed distribution.}
\componentfield{Round~2: execute the updated plan}{The proposed Actor reads the tracker through \texttt{values:batchGet}, combines the distribution criteria with the legal restrictions, and sends the release to three eligible outlets plus a summary to the team.}
\componentfield{Round~2: inspect delivered effects}{The proposed Evaluator retrieves one release and the summary with \texttt{messages.get}, \texttt{format=full}. The returned recipients are \texttt{press@techwire.example.com} and \texttt{pr-team@company.example.com}; both messages have \texttt{label\_ids=[SENT]}. Their bodies expose the release and the distribution decisions for inspection. These two reads verify the sampled messages; the other sends are recorded in the Actor trace.}
\componentfield{Adoption and outcome}{Both role updates are adopted. The selected Round~2 output passes native offline scoring and is retained in Round~3.}
\end{harnessbox}

\endgroup
\begingroup
\renewcommand{\componentfield}[2]{\par\noindent{\rmfamily\bfseries\color{harnessblue}#1.}\enspace #2\par}

\begin{harnessbox}{Delivered Content: Evolved Evaluator Rubric}{Complete recorded criterion; adopted in Round~3 with corresponding prompt and skill edits}
\label{lst:content-rubric}
\componentfield{Criterion}{Delivered message and record content compliance}
\componentfield{Verification plan}{Inspect the actual delivered message text (e.g., Slack postMessage body) and record content against the public instruction to list only items acted on and handle exclusions silently. Distinguish a required findings report (what was found) from narration of items skipped or excluded. Bind the content to the exact candidate and message.}
\componentfield{Falsifier}{A delivered message or record names, enumerates, or explains items the candidate skipped, excluded, or rejected, or lists items the candidate did not act on, contrary to the public instruction to handle exclusions silently.}
\componentfield{Unknown rule}{If the instruction's applicability to a particular message is ambiguous (e.g., a required findings report vs. an optional summary), state the competing interpretation rather than assuming \mbox{a violation.}}
\end{harnessbox}

\begin{harnessbox}{Cross-channel Deduplication: Revisiting the Same Pair}{Condensed execution record; quoted message text preserved}
\label{lst:content-execution}
\componentfield{Round~2: apparent equivalence}{Candidate-specific \texttt{tag=duplicate} reads cover four duplicate conversations in each saved state; matched records establish four log entries and a Slack notification. Preference reports no supported difference and retains the initial output. Archived versus resolved status is not treated as a defect because the public task does not prescribe either one.}
\componentfield{Round~3: inspect the missing dimension}{The Round~3 verification calls \path{conversations.history} separately on each candidate's saved state, with \texttt{channel=C\_DEDUP} and \texttt{limit=10}. Both returns have \texttt{has\_more=false}. The initial output's delivered message includes ``No other cross-channel duplicates found; independent issues were left untouched.'' The Round~2 output lists only the four pairs acted on. The former violates the public instruction to handle exclusions silently.}
\componentfield{Updated comparison}{With the same two candidate IDs, preference changes from equivalence to a strict preference for the Round~2 output in both presentation orders. The round also records Evaluator adoption and no Actor adoption. The qualification candidate IDs are not specified in this summary.}
\componentfield{Retained outcome}{The selected Round~2 output passes native offline scoring. Both fresh Round~3 attempts fail, so the successful final result comes from selecting an existing output through improved assessment.}
\end{harnessbox}

\endgroup
\begin{harnessbox}{Actor Skill: Response Scope}{Loaded and adopted in Round 2}
\label{lst:actor-skill}
\componentfield{When to use}{The task involves selecting a subset of subjects (e.g., eligible recipients) and the final response or records might narrate which subjects were skipped, excluded, or rejected.}
\componentfield{Procedure}{Before finalizing any message or record, identify the public output-scope instruction (e.g., 'list only items you acted on; do not name, enumerate, or explain items you skipped, excluded, or rejected'). Review the draft for sentences that name, enumerate, or explain skipped, excluded, or rejected items. Remove or rephrase them so exclusions are handled silently in the action. If an unfulfilled obligation must be disclosed (e.g., a required write could not be performed), state the obligation and the observed limitation without naming or enumerating the specific \mbox{skipped subjects.} If the user explicitly requested a list of exclusions or a full summary, follow that more specific instruction. Keep internal verification separate \mbox{from the final response.}}
\end{harnessbox}

\begin{harnessbox}{One Update, Two Working Checks}{Verbatim check and counterexample fields from the Sales063 proposal}
\label{lst:paired-working-checks}
\componentfield{Actor check}{Does the proposal Actor's final response comply with the output-scope rule by listing only acted-on items and not naming, enumerating, or explaining skipped/excluded/rejected attendees?}
\componentfield{Actor counterexample}{The final response contains a sentence like 'Dropped-off early attendees were not contacted' or names an excluded attendee; that would violate the instruction.}
\componentfield{Evaluator check}{Does the proposal Evaluator's report detect and flag any final-response narration of skipped/excluded/rejected items as a violation of the output-scope rule, citing the exact sentence and the public instruction?}
\componentfield{Evaluator counterexample}{The evaluator report says 'both correctly skipped early drop-offs' without noting the narration violation, or treats the narration as a required disclosure.}
\end{harnessbox}
\input{sections/appendix_deliverable_harness}
\begin{harnessbox}{Evaluator Skill: Source Breakdown}{Adopted in the tour-budget episode}
\label{lst:budget-skill}
When the task requires a breakdown by source (Tour Manager vs. production company) with a combined total, verify that each expense and revenue line is correctly attributed to the source columns and that the combined column equals the sum of the two source columns. Do not rely solely on category labels; check the actual numeric values in each column against the reference files.
\end{harnessbox}
\begin{harnessbox}{Evaluator Skill: Date and Spending Rules}{Adopted in the grant-spending episode}
\label{lst:grant-skill}
When a task requires date-based percentage calculations, verify the exact date arithmetic (e.g., inclusive vs exclusive day counts, as-of date handling) against the task's stated criteria and boundary conditions. Confirm that the computed percentages and resulting classifications (e.g., Fast/Slow Spending) match the task's explicit thresholds, and test at least one boundary case to ensure the comparison operators are \mbox{correctly applied.}
\end{harnessbox}
\input{sections/appendix_python_checks}
\input{sections/appendix_environment_checks}

\begin{harnessbox}{Evidence Attribution Instructions}{Fixed prompt excerpts; enabled with comparative claims}
\label{lst:evidence-instructions}
When finishing, return comparative\_claims. Bind every claim to candidate\_ids. Use claim\_ref candidate:<candidate\_id> for a candidate statement or action, or task\_context for a shared task requirement. This exact frozen-source reference locates the claim but does not verify it. Use evidence\_ref task\_context only for separate public task or policy evidence, or the public\_evidence\_ref returned by an executed tool. Use impact supports\_candidate or harms\_candidate only for a single candidate when the verified fact changes task quality; shared or nonconsequential facts are neutral. If neither source supports or contradicts the claim, use status unknown and evidence\_ref none. Claims are not winners or scores.

For each comparative claim distinguish the observed effect from the public rule and the premise that makes the rule applicable to this candidate. Record rule\_support and applicability\_premise with their own evidence and uncertainty. A delivered output alone does not verify its correctness. If no rule or connecting premise is established, state that uncertainty explicitly; do not invent one. These are fallible assessments for Preference, not admission thresholds.
\end{harnessbox}

\begin{harnessbox}{Joint Update Proposal: Response Scope}{Recorded proposal}
\label{lst:joint-hypothesis}
\componentfield{Observed gap}{The current evaluator missed a shared violation of the public output-scope instruction: both candidates' final responses narrated skipped/excluded attendees (e.g., 'Dropped-off early attendees were not contacted', 'attendees below the 45-minute threshold ... were silently excluded') even though the public system instruction requires listing only acted-on items and handling exclusions silently. The evaluator instead praised the exclusions without checking the narration rule, so the internal signal can fail to detect a clear public-instruction violation.}
\componentfield{Coupled change}{Add an evaluator rubric criterion and skill that explicitly check final responses (and records) against the public output-scope rule, and add an Actor skill that reinforces silent exclusion handling before finalizing. The bundle aligns both roles on the same public instruction: the Actor avoids narrating exclusions, and the Evaluator flags any narration that remains.}
\componentfield{Expected observation}{In a fresh pair, the proposal Actor's final response lists only acted-on items (sent emails, CRM limitation) and does not name or explain excluded attendees; the proposal Evaluator's report flags any candidate that narrates exclusions as violating the output-scope rule, citing the exact sentence and the public instruction.}
\componentfield{Counterexample}{A task whose user explicitly requests a summary of excluded recipients or a full exclusion list: narration is then required, and the Evaluator must not flag it. Also, a generic disclosure of an unfulfilled obligation (e.g., 'CRM could not be updated') is not prohibited narration unless it names/enumerates the \mbox{skipped subjects.}}
\end{harnessbox}
\begin{harnessbox}{Date and Threshold Check}{Recorded executed program}
\label{lst:grant-program}
\begin{lstlisting}[basicstyle=\ttfamily\small,breaklines=true,breakatwhitespace=false,columns=fullflexible,keepspaces=true]
from datetime import date
start = date(2024, 10, 1)
end = date(2027, 9, 30)
asof = date(2025, 3, 31)
total = (end - start).days
elapsed = (asof - start).days
pct = elapsed / total * 100
print(f'elapsed_days={elapsed}, total_days={total}, pct={pct:.2f}')
print(f'funds_pct=99.63, fast={99.63 > 50 and pct <= 25}')
\end{lstlisting}
\componentfield{Returned observation}{181 elapsed days; 1,094 total days; 16.54\% elapsed; 99.63\% of funds spent; the fast-spending condition evaluates to true.}
\end{harnessbox}

%% file: sections/appendix_deliverable_harness.tex
\begin{harnessbox}{GDPval Actor: Produce and Reopen}{Recorded initialization policy, rubric, and skill}
\label{lst:gdp-actor}
\componentfield{Instruction}{Translate the public task into a scoped, usable deliverable; preserve required facts and conditions. Verify the exact object being changed and then reload the produced artifact to inspect the requested effect. Satisfy the task in a practical way rather than exploiting a minimal literal interpretation.}
\componentfield{Working rubric}{Bind every requested change to its intended file, entity, sheet/section, units, dates and conditions; verify the produced artifact preserves unaffected facts and achieves the intended use.}
\componentfield{Skill}{Keep rule applicability: distinguish explicit requirements, necessary implications, optional improvements and unresolved assumptions; preserve alternatives and exceptions. Treat extra detail as optional unless it serves the requested use. Before editing, inspect real template structure, styles, worksheet names and target rows. After writing, use the preinstalled native probe on output/ (not merely reference\_files/ or program source), check the relevant content/structure and intended effect, and preserve before/after evidence. Import only needed libraries; use simple requested formats rather than inventing extra deliverables.}
\end{harnessbox}

\begin{harnessbox}{GDPval Evaluator: Inspect and Refine}{Recorded initialization policy, rubric, and skill}
\label{lst:gdp-evaluator}
\componentfield{Instruction}{Improve and reuse the current verification procedure instead of accumulating criteria. Begin with the unresolved decision-relevant question, its public basis and applicability conditions; inspect actual candidate files. Distinguish a usable allowed solution from both gratuitous additions and merely nominal compliance.}
\componentfield{Working rubric}{A criterion must serve the actual task or a justified necessary implication; preserve scope, alternatives, exceptions and approximate-versus-exact modality.

Compare operational completeness and supported meaning using observed candidate evidence; evidence validity, counterexamples and uncertainty matter more than confidence, length or repeated checks.}
\componentfield{Skill}{Reuse an existing check if its target, conditions and evidence coverage still apply. Do not add another check with the same target and evidence scope; merge or repair the existing check. Prefer native structure/literal/table probes to long generated programs. Bind each observation to the actual candidate object and its coverage; a reference/template hit is not proof of generated content, a witness is not exhaustive coverage, and an incomplete or failed search cannot establish absence. Correct the smallest defective check and retain a known success and a counterexample. Describe shared fulfillment, actual differences and unresolved premises. A report supplies fallible evidence; it must not prescribe the selector winner or claim that uncertain facts establish equivalence. For a requested clickable workbook link, inspect native hyperlink targets with the hyperlinks probe; URL text alone is insufficient, and a stored target does not prove destination availability.}
\end{harnessbox}

%% file: sections/appendix_python_checks.tex
\lstdefinestyle{vbdpython}{%
  language=Python,
  basicstyle=\fontsize{9}{10.5}\selectfont\ttfamily,
  keywordstyle=\color{harnessblue}\bfseries,
  commentstyle=\color{gray},
  stringstyle=\color{harnessblue!80!black},
  showstringspaces=false,
  columns=fullflexible,
  keepspaces=true,
  breaklines=true,
  breakatwhitespace=false,
  breakindent=1em,
  emptylines=0,
  tabsize=4,
  aboveskip=5pt,
  belowskip=5pt
}

\begin{harnessbox}{Award Population Reconciliation}{Executed Python; source workbook and two candidate workbooks}
\label{lst:award-population-check}
\componentfield{Verification target}{Recompute which awards qualify for the requested spending analysis, then check whether both candidate workbooks contain exactly those award numbers. The calculation combines date intervals, expenditures, award amounts, and two threshold rules.}
\begin{lstlisting}[style=vbdpython]
import openpyxl, glob

src = glob.glob("reference_files/*.xlsx")[0]
wb = openpyxl.load_workbook(src, data_only=True)
ws = wb["Award Data"]
rows = list(ws.iter_rows(values_only=True))
header = rows[0]
si = {h: i for (i, h) in enumerate(header)}
from datetime import date

asof = date(2025, 3, 31)
qual = []
for r in rows[1:]:
    if r[si["Start Date"]] is None or r[si["End Date"]] is None:
        continue
    start = (r[si["Start Date"]].date()
             if hasattr(r[si["Start Date"]], "date")
             else r[si["Start Date"]])
    end = (r[si["End Date"]].date()
           if hasattr(r[si["End Date"]], "date")
           else r[si["End Date"]])
    total = r[si["Total Awarded Amt"]]
    ffr = r[si["FFR Expenditure Amt"]]
    if total is None or ffr is None or total == 0:
        continue
    pct_time = (asof - start).days / (end - start).days * 100
    pct_spent = ffr / total * 100
    if pct_time <= 25 and pct_spent > 50:
        qual.append((r[si["Recipient Award Number"]], "Fast"))
    elif pct_time >= 75 and pct_spent < 25:
        qual.append((r[si["Recipient Award Number"]], "Slow"))
print("Qualified count:", len(qual))
print("Qualified IDs:", sorted((x[0] for x in qual)))
for cid in ["draft-r0-0", "draft-r0-1"]:
    out = glob.glob(f"candidates/{cid}/output/*.xlsx")[0]
    wb2 = openpyxl.load_workbook(out, data_only=True)
    ws2 = wb2["Spending Rate Analysis"]
    rows2 = list(ws2.iter_rows(values_only=True))
    ids = [r[0] for r in rows2[1:]]
    print(cid, "IDs match:", sorted(ids) == sorted((x[0] for x in qual)),
          "count:", len(ids))
    for r in rows2[1:4]:
        print(cid, r[0], r[3], r[6], r[7])
\end{lstlisting}
\componentfield{Returned observation}{The source calculation produces 53 qualifying awards. Both candidate ID lists match exactly. Three rows' percentage values are printed for inspection, showing decimal fractions in one output and percentage notation in the other.}
\end{harnessbox}

\begin{harnessbox}{Native Pivot Table Inspection}{Executed Python; inspection of the XLSX package}
\label{lst:pivot-structure-check}
\componentfield{Verification target}{Determine whether the generated workbook contains actual pivot tables for the requested sales summaries. A sheet displaying totals can be a static table, so the check inspects the workbook package rather than relying on displayed values.}
\begin{lstlisting}[style=vbdpython]
import zipfile
import xml.etree.ElementTree as ET

path = "candidates/draft-r0-0/output/Weekly Sales Data.xlsx"
with zipfile.ZipFile(path) as z:
    names = z.namelist()
    print("PIVOT CACHE FILES:", [n for n in names if "pivotCache" in n])
    print("PIVOT TABLE FILES:", [n for n in names if "pivotTable" in n])
    wb = z.read("xl/workbook.xml").decode("utf-8")
    print("WORKBOOK HAS PIVOT CACHE REF:", "pivotCaches" in wb)
    for n in names:
        if n.startswith("xl/worksheets/sheet") and n.endswith(".xml"):
            data = z.read(n).decode("utf-8")
            if "pivotTable" in data:
                print("SHEET", n, "HAS PIVOT TABLE REF")
    for n in names:
        if "pivotTable" in n and n.endswith(".xml"):
            print("---", n, "---")
            print(z.read(n).decode("utf-8")[:2000])
\end{lstlisting}
\componentfield{Returned observation}{\texttt{PIVOT CACHE FILES: []}; \texttt{PIVOT TABLE FILES: []}; \texttt{WORKBOOK}\allowbreak\ \texttt{HAS}\allowbreak\ \texttt{PIVOT}\allowbreak\ \texttt{CACHE}\allowbreak\ \texttt{REF:}\allowbreak\ \texttt{False}.}
\end{harnessbox}
\begin{harnessbox}{Cumulative Capacity Verification}{Executed Python; recomputation across thirteen weekly records}
\label{lst:capacity-balance-check}
\componentfield{Verification target}{Check whether the catch-up plan's cumulative balance follows weekly capacity minus demand, starting from a backlog of 438.81 hours, and identify when the balance first becomes nonnegative.}
\begin{lstlisting}[style=vbdpython]
import openpyxl

wb = openpyxl.load_workbook(
    "candidates/draft-r0-0/output/Catch-Up_Plan_MIG_Welding.xlsx"
)
ws = wb["Catch-Up Plan"]
weeks = []
for r in range(5, 18):
    week = ws.cell(row=r, column=1).value
    demand = ws.cell(row=r, column=2).value
    capacity = ws.cell(row=r, column=5).value
    cum_bal = ws.cell(row=r, column=7).value
    weeks.append((week, demand, capacity, cum_bal))
print("Weeks 4-16 (week, demand, capacity, cum_bal):")
for w in weeks:
    print(w)
bal = -438.81
for week, demand, capacity, cum_bal in weeks:
    bal = bal + capacity - demand
    print(f"Week {week}: calculated balance {bal:.6f}, stored {cum_bal}")
\end{lstlisting}
\componentfield{Returned observation}{The recomputed and stored balances agree to the displayed precision over weeks 4--16. Week 13 has a balance of \textminus21.422573 hours; week 14 has 19.131123 hours, the first nonnegative balance in the inspected schedule.}
\end{harnessbox}

%% file: sections/appendix_environment_checks.tex
\begin{harnessbox}{Policy Availability and Delivered Content}{Sales052; Round~3 review of current and proposed Evaluators on the same candidates}
\label{lst:ab-policy-content}
\componentfield{Verification target}{The task requests an escalation based on a risk scoring framework. Both new candidates derive score 8 and perform the core escalation actions. A policy email additionally requests the tracking reference \texttt{ESC-2026-Q1}. Before judging its omission, the Evaluator must establish that the policy was available to the candidate and inspect what that candidate actually delivered.}
\componentfield{Recorded skill procedure}{When a candidate omits a field claimed to be required by a business policy or instruction record, first verify that the policy source exists and is readable in that candidate's own world (e.g., read the email/document in that candidate's after-state). Treat an omission as a deficiency only when the source was available but unread; if availability cannot be verified, keep the omission as an unresolved interpretation. Label the policy's binding status as an interpretation unless the public task text itself names the requirement.}
\componentfield{Executed requests}{The proposed Evaluator performs both reads on candidate \texttt{fe24fe00\ldots}'s own saved state. The current Evaluator reads the same email but does not independently retrieve the Slack message.}
\begin{lstlisting}[style=vbdpython,language={}]
GET https://gmail.googleapis.com/gmail/v1/users/me/messages/msg_esc_policy_001
params: {"format": "full"}

GET https://slack.com/api/conversations.history
params: {"channel": "C004LEADERSHIP", "limit": 50}
\end{lstlisting}
\componentfield{Returned observations}{Both reads are complete. The email says, ``All deal escalation posts must include the tracking reference ESC-2026-Q1 for compliance audit trail.'' The Slack return contains one message, timestamp \texttt{1790310961.949455}, with no continuation cursor. It includes the score \texttt{8/8}, the amount \texttt{\$500,000 annual}, and the calculation \texttt{3 + 2 + 2 + 1 = 8}, but omits the tracking reference.}
\componentfield{Comparative review}{Both Evaluators identify the omission. The reviewer favors the proposed procedure in both presentation orders because it directly inspects the disputed message and distinguishes that observation from the other candidate's write acknowledgment. The reviewer notes that the current report claims broader message readback than its executed checks support.}
\componentfield{Decision and scope}{The Evaluator update is adopted; the previously selected Round~2 output is retained, and the proposed Actor is not adopted. The report treats the policy email's binding status as an interpretation because the task does not name the tracking reference.}
\end{harnessbox}

\begin{harnessbox}{Recent Updates and Journal Content}{Finance097; Round~2 evidence followed by an Evaluator skill adopted in Round~3}
\label{lst:ab-recent-updates}
\componentfield{Verification target}{The user requests a revenue recognition journal and warns that contract statuses may have changed. One candidate follows a schedule marking Meridian's milestone complete and sends a journal totaling \$80,000. Another reads a Slack update that the milestone is incomplete and sends a \$40,000 journal. The Evaluator tests whether the first candidate could also access the update.}
\componentfield{Executed request in Round~2}{This read is bound to candidate \texttt{79fc3075\ldots}, whose journal includes the disputed recognition.}
\begin{lstlisting}[style=vbdpython,language={}]
GET https://slack.com/api/conversations.history
params: {"channel": "C_REV", "limit": "50"}
\end{lstlisting}
\componentfield{Returned observation}{The complete response contains a message by \texttt{U\_REV\_MGR}, timestamp \path{1769904000.000100}: ``Update: the Meridian Implementation Phase 1 milestone is NOT complete yet. QA found issues and pushed it to March. Do not recognize the milestone revenue this month.'' The observation establishes that the update exists in the candidate's own environment. Its Actor trajectory contains no Slack query.}
\componentfield{Evidence about the delivered journal}{The comparison reuses an earlier qualification read on the other candidate, \texttt{43e68118\ldots}:}
\begin{lstlisting}[style=vbdpython,language={}]
GET https://gmail.googleapis.com/gmail/v1/users/me/messages/41723a4088914c99
params: {"format": "full"}
\end{lstlisting}
The complete return has label \texttt{SENT}, recipient \path{controller@company.example.com}, and journal entries of \$10,000, \$5,000, and \$25,000, totaling \$40,000. The Round~2 final comparison uses one new Slack read and this reused email read; the \$80,000 journal is supported by its own Actor's recorded readback.
\componentfield{Recorded skill procedure adopted in Round~3}{Identify the plausible channels where the flagged information could live (email, Drive, Slack, and other messaging or collaboration systems). For each candidate, check the trajectory for whether it queried those channels. If a candidate did not query a channel, use a permitted read on that candidate's own snapshot to determine whether an in-scope update exists there. If an available update exists and the candidate's deliverable does not reflect it, treat the difference in the delivered deliverable content as material rather than an optional extra, while keeping the relative authority of conflicting sources (e.g., a stale schedule marker versus a newer update) explicitly unresolved. Bind the read to the exact candidate and snapshot; do not transfer it to the other candidate.}
\componentfield{Decision and scope}{Round~2 selects the \$40,000 journal, which passes offline scoring. The observation precedes the skill's adoption in Round~3. The report keeps the relative authority of the Slack update and the schedule explicit as an unresolved interpretation, and does not rank candidates merely because one additionally updates the source schedule.}
\end{harnessbox}